\documentclass[10pt,twocolumn,letterpaper]{article}

\PassOptionsToPackage{table}{xcolor}
\usepackage[pagenumbers]{iccv} 

\usepackage{amsmath,amssymb}
\usepackage{multirow}

\usepackage[mathscr]{euscript}
\usepackage{booktabs}

\usepackage{xspace}
\newcommand{\pbest}{\ensuremath{P_{\text{Best}}}\xspace}
\newcommand{\methodname}{\textup{CODA}\xspace}

\makeatletter
\def\and{%
  \end{tabular}%
  \hskip 0.9em 
  \begin{tabular}[t]{c}}
\makeatother

\definecolor{iccvblue}{rgb}{0.21,0.49,0.74}
\usepackage[pagebackref,breaklinks,colorlinks,allcolors=iccvblue]{hyperref}

\def\paperID{9308} 
\def\confName{ICCV}
\def\confYear{2025}

\title{Consensus-Driven Active Model Selection}

\author{Justin~Kay\thanks{Correspondence to: {\tt\small kayj@mit.edu}}\\
MIT \\
\and
Grant~Van~Horn\\
UMass Amherst \and 
Subhransu~Maji \\
UMass Amherst\and 
Daniel Sheldon \\
UMass Amherst\and 
Sara Beery\\
MIT
}

\begin{document}
\maketitle

\begin{abstract}

The widespread availability of off-the-shelf machine learning models poses a challenge: which model, of the many available candidates, should be chosen for a given data analysis task?
This question of \textup{model selection} is traditionally answered by collecting and annotating a validation dataset---a costly and time-intensive process.
We propose a method for \textup{active} model selection, using predictions from candidate models to prioritize the labeling of test data points that efficiently differentiate the best candidate. 
Our method, \textup{\textbf{CODA}}, performs \textbf{co}nsensus-\textbf{d}riven \textbf{a}ctive model selection by modeling relationships between classifiers, categories, and data points within a probabilistic framework. 
The framework uses the consensus and disagreement between models in the candidate pool to guide the label acquisition process, and Bayesian inference to update beliefs about which model is best as more information is collected.
We validate our approach by curating a collection of 26 benchmark tasks capturing a range of model selection scenarios.
\methodname outperforms existing methods for active model selection significantly, reducing the annotation effort required to discover the best model by upwards of 70\% compared to the previous state-of-the-art. Code and data are available at: \href{https://github.com/justinkay/coda}{https://github.com/justinkay/coda}.

\end{abstract}

\vspace{-10pt}
\section{Introduction}

\begin{figure}
    \centering
    \includegraphics[width=0.9\linewidth]{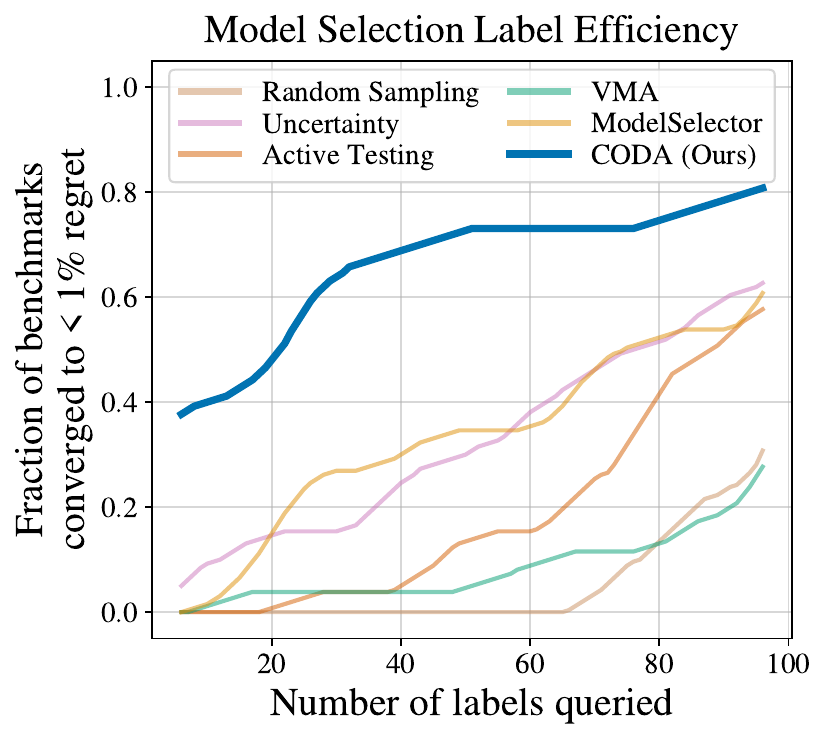}
    \vspace{-10pt}
    \caption{\textbf{We introduce CODA, a consensus-driven method for active model selection.} This figure shows the number of labels needed to converge to the optimal or near-optimal (within 1\% accuracy) model in a benchmark suite of 26 model selection tasks. CODA is significantly more label-efficient than prior work, identifying a near-optimal model with fewer than 25 labeled examples over 50\% of the time, and with fewer than 100 labeled examples over 80\% of the time.}
    \label{fig:fig1}
    \vspace{-5pt}
\end{figure}

The availability of off-the-shelf machine learning models is growing rapidly. 
As of this writing there are over 1.9M pre-trained models available for download from the HuggingFace Models repository~\cite{huggingface}, ranging from small specialized models to large general-purpose foundation models. Application-specific \textit{model zoos} are growing as well, curating sets of models for everything from wildlife monitoring~\cite{hernandez2024pytorch} to medicine~\cite{cardoso2022monai,ouyang2022bioimage}.
These zoos potentially enable accurate data analysis without the need for custom ML development, but introduce a new challenge: \textbf{which model, of the many available, performs best for a given set of data?} Traditionally \textit{model selection} decisions are made by collecting labels for a subset of the data in question and evaluating the performance of each model on that subset. To ensure results are robust, these datasets need to be large, representing significant human effort for each new dataset.

While reducing human effort during \textit{training} has been well-studied~\cite{li2024survey,settles2009active}, efficient model selection at test time is relatively unexplored. Progress on this challenge will be beneficial for both users of pre-trained models and for researchers designing label-efficient algorithms. In particular, the field of \textit{unsupervised domain adaptation (UDA)} proposes to adapt algorithms to new data without any human labels whatsoever, yet successful UDA methods are highly dependent on the use of human-labeled validation sets for model selection~\cite{musgrave2021unsupervised,ericsson2023better,kalluri2024uda,kay2024align,kay2023unsupervised}. This contradiction has motivated work in \textit{unsupervised model selection}~\cite{musgrave2022three,you2019towards,saito2021tune}, but so far these methods have proven to be unreliable, especially in challenging real-world conditions~\cite{musgrave2021unsupervised,kay2023unsupervised,ericsson2023better,kalluri2024uda}.
 
Recently, methods for \textit{active model selection} have been proposed to identify an optimal model from a candidate set with fewer labels than required by traditional fully-labeled validation~\cite{matsuura2023active,sawade2012active,okanovic2024all,karimi2021online}. Active methods use model predictions to guide the label acquisition process, iteratively querying a human expert for labels on specific data points that are expected to be most informative. 
Though promising, prior work remains label-inefficient, often requiring several hundred to several thousand labels to reliably perform model selection~\cite{okanovic2024all,matsuura2023active,sawade2012active}. There are two key limitations that lead to this inefficiency: (1) models are largely treated independently of each other, both before and during the label collection process, ignoring valuable information captured by model agreement and disagreement; (2)~categories are also treated independently, ignoring correlations between data points that can be deduced from category-specific model errors.

In this paper we propose a novel \textbf{co}nsensus-\textbf{d}riven \textbf{a}ctive model selection method, \textbf{CODA}, to address these limitations. Our approach models relationships between classifiers, categories, and data points in order to make more informed label queries. To do this we revisit classical probabilistic models of the classification data generating process. 
Specifically, we propose a framework inspired by the Dawid and Skene model of annotator agreement~\cite{dawid1979maximum,passonneau2014benefits}, whereby each classifier is represented by a \textit{confusion matrix} that captures its per-category performance characteristics. We adapt this framework for active model selection by constructing a probabilistic estimate over which model is best that accounts for per-category classifier consensus and uncertainty. We then iteratively query for ground-truth labels on data points that are expected to provide maximal information about the probability that each model is the best.

We validate our approach by curating a benchmarking suite of 26 model selection tasks representing a variety of real-world use cases across computer vision and natural language processing, which we publish alongside our method to support future model selection research. Our approach exceeds, often significantly, the performance of the previous state-of-the-art on 18 out of these 26 tasks. In addition, our method is exceptionally label-efficient, often requiring fewer than 25 labeled examples to identify the best or near-best model (see \cref{fig:fig1}). 

In summary, our main contributions are the following:

\begin{enumerate}
    \item We introduce CODA, a novel method for active model selection. CODA leverages model consensuses and Bayesian inference to identify the most informative labels for performing model selection at test time.
    \item We curate a benchmarking suite of 26 active model selection tasks to validate our approach and compare with prior work. We release the data publicly to support future research in active model selection.
    \item We demonstrate that CODA achieves state-of-the-art performance on 18 out of 26 tasks in our benchmark. Additionally, though not the main focus of our work, we show that CODA's initialization routine allows us to match or exceed state-of-the-art \textit{unsupervised} model selection results on 20 out of 26 tasks.
    
\end{enumerate}

\section{Related work}

\noindent \textbf{Model selection in machine learning} is typically performed using a held-out ``validation'' set to select between different candidate algorithms, hyperparameters, and/or training checkpoints.
Differences between training, validation, and test distributions create challenges for model selection~\cite{koh2021wilds,beery2018recognition,kay2022caltech}.
While in some cases out-of-distribution accuracy has been observed to be highly linearly correlated with in-distribution accuracy~\cite{miller2021accuracy,recht2019imagenet,recht2018cifar}, in others it has been observed to be uncorrelated or even negatively correlated~\cite{miller2021accuracy,teney2023id,sanyal2024accuracy}, indicating that reliable model selection for a dataset of interest cannot in general be performed using a validation set from a different data distribution. This has implications both for pre-trained models sourced from model zoos, where little or nothing is known about the training data distribution, as well as for models trained on one distribution and deployed on another. Our experiments and benchmarking suite evaluate both settings.

\noindent \textbf{Unsupervised model selection} methods perform model selection without the use of test labels. These methods have largely been proposed in the context of \textit{unsupervised domain adaptation}, where \textit{unlabeled} test-domain data is available for training. Unsupervised validation methods in this setting typically compute a heuristic such as entropy based on model predictions on this unlabeled test-domain data, under the assumption that these measures are correlated with accuracy~\cite{musgrave2022three,yang2023can,robbiano2022adversarial,saito2021tune}. 
Alternatively, methods may use the accuracy on labeled in-distribution examples, weighted by their ``similarity'' to out-of-distribution samples, as a proxy~\cite{you2019towards}. Unfortunately, many of these methods have been shown to be poorly or even negatively correlated with test-domain accuracy~\cite{musgrave2021unsupervised,ericsson2023better,kalluri2024uda}. 

One key limitation of this family of methods is that they consider the predictions of individual model checkpoints in isolation from any other models being evaluated; in contrast, recent work has identified the possibility to utilize the predictions of all models concurrently to better estimate their individual performance~\cite{hu2025towards,shanmugam2025evaluating}. Our method also harnesses this consensus information, but uses a probabilistic framework to aggregate and update our beliefs about the prediction set over time.

\noindent \textbf{Active learning and active testing}
methods intelligently select informative data points to reduce annotation effort for training and evaluating machine learning models~\cite{li2024survey,kossen2021active,kossen2022active,nguyen2018active,settles2009active}. 
Active testing is related to our setup but differs in that the goal is to estimate the test loss of one model, rather than select the best from a set of candidates.
Existing methods function by constructing unbiased importance sampling estimators of model performance~\cite{kossen2021active}. While it is possible to perform active model selection by performing active testing concurrently for all models and selecting the one with the lowest loss estimate, we will demonstrate this is significantly less label-efficient than specific targeted strategies for active model selection.

\noindent \textbf{Active model selection}
methods have focused predominantly on the \textit{online} setting, where data points are observed in a stream~\cite{karimi2021online,kassraie2023anytimemodelselectionlinear,liu2022contextual}. In contrast, we focus on the \textit{pool-based} setting where a static collection of unlabeled data is available from the beginning. 
Early work in the pool-based setting resembles work in active testing, using importance-weighted loss estimates for each model~\cite{sawade2012active,matsuura2023active}. As pointed out by \citet{kossen2022active}, these importance-sampling-based approaches exhibit high variance early in the sampling process, since metrics are computed solely from the labels collected so far.
Practically, this means model selection remains unreliable until a significant number of annotations have been collected. 

Recent work from \citet{okanovic2024all} performs active model selection without importance sampling estimators by defining a simple single-parameter distribution over which model is best at any given time. At each time step, the posterior probability for each model being best is updated according to $\text{Posterior} = \frac{1-\epsilon}{\epsilon} \times \text{Prior}$ if the model gets the label correct, where $\epsilon$ is a hyperparameter determining the learning rate. This simple probabilistic approach has been shown to be more label-efficient than prior work, but still requires significant annotation effort to overcome both its uninformative priors and independence assumptions between data points. Our method addresses these limitations by constructing informative unsupervised priors and by modeling correlated errors across the test data pool.

\noindent \textbf{Probabilistic models of agreement} aggregate annotations created by a group of human annotators on a dataset. They do so by modeling a ``data generating process'' that describes how annotations are created according to latent random variables like per-annotator accuracy. The Dawid-Skene model~\cite{dawid1979maximum} (which we describe in more detail in \cref{sec:ds}) is an early example that remains popular today. Initially proposed for aggregating the predictions of doctors regarding patient outcomes, it has since found success in cleaning crowd-sourced annotations~\cite{ratner2017snorkel,welinder2010multidimensional} and merging human- and AI-generated predictions~\cite{branson2017lean,van2018lean,tamura2024influence}. Many extensions have been proposed that incorporate Bayesian inference~\cite{passonneau2014benefits,paun2018comparing} or more complex data generating processes~\cite{braylan2023general,welinder2010multidimensional,hovy2013learning}. We extend the general latent variable framework for active model selection. We do not fit the model directly to predictions as in prior work; instead, we use the framework as a starting point and update it iteratively to incorporate actively-collected information.

\section{Active model selection problem formulation} 

\noindent \textbf{Models and data} 
We assume that we have a hypothesis set \( H = \{ h_k \}_{k=1}^{|H|}\) consisting of candidate models from which we want to select. Each model has generated predictions on some unlabeled test set \( D = \{ x_i \}_{i=1}^{|D|} \) that we care about but cannot exhaustively annotate. We assume the predictive task is a $C$-way multi-class classification problem with 
    $\hat{c}_{k,i} = \arg\max_c  h_k(x_i)$,
where $\hat{c}_{k,i} \in \{1, \ldots, C\}$ and $h_k(x_i) \in [0,1]^C$ are the class prediction and the $C$-dimensional prediction vector of model $h_k$ on data point $x_i$, respectively. Our setup is agnostic to whether $h_k(x_i)$ is a soft score vector or one-hot labels.

\noindent \textbf{Active data point selection} 
At every time step $t$, we query a human expert for a single ground truth label \( y_i \). The choice of which \( y_i \) to query for is up to the model selection algorithm.
We partition \( D \) into disjoint unlabeled and labeled subsets: \( D = D_U \cup D_L \). 
Once a point has been queried for a ground-truth label it moves from  \( D_U \) to \( D_L \). 

\noindent \textbf{Model selection and evaluation} At each time step $t$, a model selection algorithm returns its choice $\hat{h}^{(t)}$ of the model it currently believes is best.
To evaluate these choices we assume that we know the form of the loss function \( L \) that we would use to evaluate each model in \( H \) if we had test labels. Our true best model, \( h^* \) (the one we hope to select), is the one that minimizes this loss empirically over \( D \):
\vspace{-5pt}
\begin{equation}
h^* = \arg\min_{h \in H} \frac{1}{|D|} \sum_{i=1}^{|D|} L\left( h(x_i), y_i \right),
\vspace{-5pt}
\end{equation}
where \( y_i \) are the true labels corresponding to \( x_i \). 
In this paper we focus on accuracy-based loss functions; extending our framework to additional metrics is an interesting direction for future work. 

We evaluate the efficacy of model selection algorithms based on the \textit{regret} incurred at each time step, defined as the difference in loss between our chosen model \( \hat{h}^{(t)} \) and the true best model \( h^* \):
\vspace{-5pt}
\begin{equation}
    \label{eq:regret}
    \text{Regret}_t = \frac{1}{|D|} \sum_{i=1}^{|D|} \left[ L\left( \hat{h}^{(t)}(x_i), y_i \right) - L\left( h^*(x_i), y_i \right) \right]
\vspace{-3pt}
\end{equation}
We also measure \textit{cumulative regret} at each time step, defined as the sum of all previous regrets:
\vspace{-5pt}
\begin{equation}
        \label{eq:cumreg}
        \text{Cuml. Regret}_t = \sum_{s=1}^{t} \text{Regret}_s
\vspace{-5pt}
    \end{equation}
Note that we can only measure these values while benchmarking as it requires oracle access to ground truth labels. 

\section{Method}

\begin{figure*}[t]
    \centering
    \includegraphics[width=\linewidth]{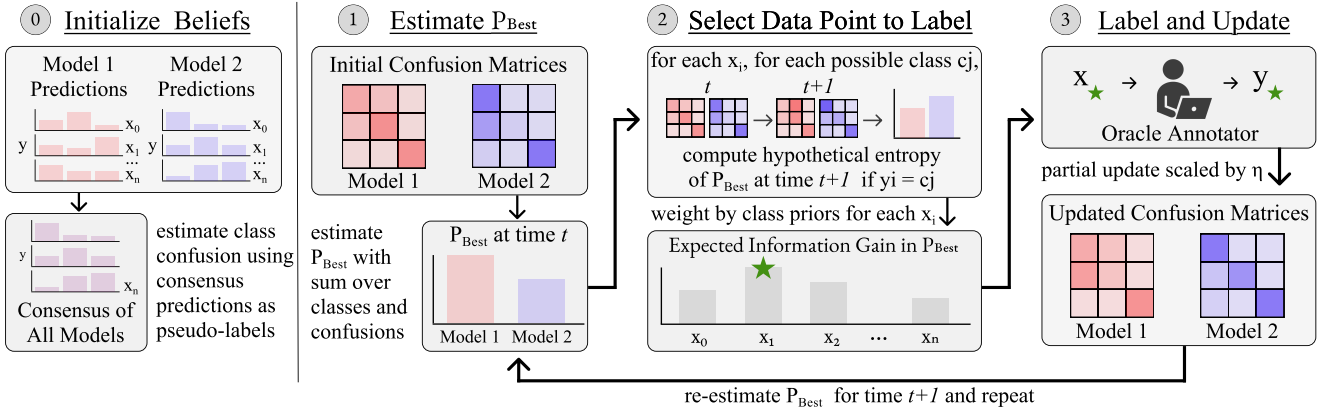}
    \vspace{-20pt}
    \caption{\textbf{\methodname for active model selection.} Simplified example with two models. At time step 0 we instantiate a Bayesian framework for tracking model performance over time (\cref{sec:ds}), using the consensus of all model predictions to instantiate per-model priors (\cref{sec:priors}). At each time step we perform three actions: \textbf{(1)}~We estimate \pbest, the current probability that each model is best, by integrating over our current beliefs (\cref{sec:pbest}); \textbf{(2)}~We compute the expected information gain in the \pbest distribution that would result from labeling each point in our dataset, and select the argmax as the most informative point (\cref{sec:eig}); \textbf{(3)}~We query for the ground truth label of our selected data point, evaluate whether each model correctly predicts the true label, and update our beliefs (\cref{sec:update}).}
    \vspace{-12pt}
    \label{fig:method}
\end{figure*}

\subsection{A Dawid-Skene model for classifier predictions}
\label{sec:ds}

The Dawid-Skene (DS) model is a probabilistic representation of how human annotators generate predictions on a set of data~\cite{dawid1979maximum}.
We adapt this model for the purpose of instead modeling the prediction process of machine learning models. Unlike prior work, we do not fit the model directly to the set of predictions; rather, we iteratively fit parameters to actively-collected ground-truth labels over time.

In particular, we base our approach off of the Bayesian instantiation of the DS model introduced by \citet{passonneau2014benefits}. We model the prediction process of each classifier $h_k$ using a confusion matrix $M_k$ of size \((C, C)\). Each row corresponds to a \emph{true class} $c \in \{1,\dots,C\}$ and each column to a \emph{predicted class} $c' \in \{1,\dots,C\}$. Thus each cell in the matrix represents the conditional probability
\begin{align}
    \label{eq:beta}
    M_{k,\, c,\, c'}
    \;=\;
    P\bigl(\hat{c}_{k,i} = c' \;\big|\; y_i = c\bigr).
\end{align}
Our goal will be to perform statistical inference on the parameters of this confusion matrix. As such, the components of the data generating process are represented by random variables with latent parameters. 
The data generating process proceeds as follows:
\begin{enumerate}
    \item Each data point's true class label $y_i$ is drawn randomly from per-data-point prior distributions over which class that data point could be, $y_i \sim \text{Cat}(\pi(x_i))$. 
    
    \item Each row of the classifier's confusion matrix is drawn randomly from per-row distributions, $M_{k,\, c,\cdot} \sim \theta_{k,c}$, where $\theta_{k,c}$ is the prior distribution over what the row of the confusion matrix could be. To accommodate Bayesian updates, we initialize each $\theta_{k,c}$ to be a Dirichlet prior.
    
    \item The sampled true class indexes into the corresponding row of the classifier's confusion matrix, $M_{k,\, y_{i}}$.
    \item The classifier's prediction for that data point is sampled from the distribution over that row's cells, ${\hat c_{k, i} \sim \text{Cat}(M_{k,\, y_{i}}})$. 
\end{enumerate}
\noindent
 See the supplemental material \cref{fig:ds} for an illustrative view.

\subsection{Constructing consensus priors (\cref{fig:method}, Step 0)}
\label{sec:priors}
We begin by collecting each model’s prediction vectors over the unlabeled dataset~$D$. We take advantage of the ``wisdom of the crowd'' (of classifiers) to form initial consensus labels by summing these probabilities across all models,
\vspace{-5pt}
\[
  s_{i,c} \;=\;
  \sum_{k=1}^H h_{k,c}(x_i) \quad
  \forall\,c=1,\ldots,C,
\]
and then define the consensus label $c_{i}^* \;=\;\arg\max_{\,c}\;s_{i,c}$.
For each model~$h_k$, we then compare its predictions 
$h_k(x_i)$
against the consensus labels $c_{i}^*$ for all $i$ to initialize empirical confusion matrices:
\vspace{-5pt}
\begin{align}
  \hat{M}_{k,c,c'}
  \;=\;
  \sum_{i=1}^{|D|}
    \left(\mathbf{1}\!\bigl[c_i^* = c\bigr]
    \cdot
    h_{k,c'}(x_i)\right)
    \label{eq:consensus}
\vspace{-5pt}
\end{align}
We then use these empirical estimates to create Dirichlet priors $\theta$ over our beliefs in each row as:
\vspace{-5pt}
\begin{align}
  \theta_{k,c,c'} &= (\beta_{c,c'} + \alpha\hat{M}_{k,c,c'})~/~T, \\
  \beta_{c,c'} &=
  \begin{cases}
  1, & \text{if } c' = c, \\
  \displaystyle \frac{1}{\,C-1\,}, & \text{otherwise.}
  \end{cases}
  \label{eq:diagonal}
\vspace{-5pt}
\end{align}
Where $\alpha$ is a hyperparameter controlling a blend between our empirical estimates and a static prior $\beta$ representing 50\% macro-accuracy, and $T$ is a temperature parameter controlling the number of initial ``pseudo-counts''. We use $\alpha=0.1$ and $T=0.5$ by default for all experiments.

\subsection{Computing P$_{\text{Best}}$ (\cref{fig:method}, Step 1)}
\label{sec:pbest}

Throughout the active label collection process, we will create and update a probability distribution representing our belief in which model is best, 
\begin{equation}
\vspace{-5pt}
\pbest \;=\; \bigl(P(h = h^*)\bigr)_{h \in H}
\end{equation}
This distribution is the key component of our model selection algorithm. At each time step, we return $\arg\max_h \pbest$ as our choice of model, and (when benchmarking) evaluate our choice by computing the regret and cumulative regret of this choice at each time step.

We focus on accuracy as our target metric. In this case a simple option would be to use the posterior means of each classifier’s overall accuracy, computed by summing each row’s diagonal probability weighted by an estimated marginal class prevalence $\hat{\pi}(c)$ derived from the consensus: 
\vspace{-15pt}
\begin{align}
    \mathrm{Acc}_{\text{mean}}(h_k)
    \;=\;
    \sum_{c=1}^C\;\hat{\pi}(c)\;M_{\,k,\,c,\,c}
    \label{eq:acc}
\end{align}
\vspace{-15pt}
\begin{align}
    \label{eq:classmarginal}
    \hat{\pi}(c) 
    &=
    \frac{1}{|D||H|} \sum_{i=1}^{|D|} \sum_{k=1}^{|H|}
    \sum_{c'=1}^C h_{k,c'}(x_i) M_{k,\,c',\,c}
\end{align}

\cref{eq:acc} has the benefit of being efficient to compute but ignores the \emph{probabilistic} nature of our estimates, failing to account for differences in uncertainty between classes/classifiers.
Remember that we have access to more than just point estimates of the confusion matrices: we model our \textit{beliefs} in what these confusion matrix entries \textit{could} be based on what we have observed so far. We do this with per-row Dirichlet distributions, allowing us to incorporate additional uncertainty into accuracy estimates as follows.
First, see that the marginal distribution for the $c$th class of the $c$th row's Dirichlet distribution (\ie the diagonal entry) is a Beta distribution with parameters:
\vspace{-5pt}
\begin{align}
\label{eq:betas}
  \alpha_{k,c} = M_{k,c,c}, 
  \quad
  \beta_{k,c} = \sum_{c' \neq c}^{C} M_{k,c,c'}
  \vspace{-10pt}
\end{align}
Then, to compute \pbest, we can integrate over the mixtures of all models’ per-row Beta distributions weighted by the class marginal $\hat{\pi}(c)$ as defined in \cref{eq:classmarginal}.
Supposing each classifier $h_k$’s performance $X_k$ is drawn independently from some distribution with PDF $f_k(x)$ and CDF $F_{k}(x)$, the integral that computes the probability that $h_k$'s draw exceeds those of all others is:
\vspace{-5pt}
\begin{align}
  P_{\text{Best}}(h_k) &= P\bigl(X_k = \max_{l} X_l \bigr) \\
  &=
  \int_{0}^{1}
    f_k(x) \prod_{l \neq k}F_{l}(x)\,\mathrm{d}x,
    \label{eq:pbest}
    \vspace{-8pt}
\end{align}
Where the PDFs and CDFs are mixtures over the per-model per-row Beta distributions from \cref{eq:betas}:
\vspace{-6pt}
\begin{align}
f_k(x) &= \sum_{c=1}^C \hat{\pi}(c) f_{k, c}(x) \\
F_l(x) &= \sum_{c=1}^C \hat{\pi}(c) F_{l, c}(x)
\vspace{-10pt}
\end{align}
where $f_{k,c}(x)$ is the PDF of $\text{Beta}(\alpha_{k,c}, \beta_{k,c})$ and $F_{l,c}$ is the CDF of $\text{Beta}(\alpha_{l,c}, \beta_{l,c})$.

Intuitively, to have $X_k$ be the largest draw, we can “fix” $X_k$ at some value $x$ (with probability density $f_k(x)$), and then require that all $X_l$ for $l\neq k$ lie at or below $x$ (which happens with probability $F_l(x)$ each). See supplemental \cref{fig:pbest} for an illustrative view.
In our implementation, we discretize $[0,1]$ and approximate the above integral using a trapezoidal rule to integrate. Finally we obtain $\pbest = \left(\pbest(h_1),\,\dots,\,\pbest(h_{|H|})\right)$.

\subsection{Selecting points to label (\cref{fig:method}, Step 2)}
\label{sec:eig}

To decide which point to label next at each time step, we aim to pick the one that, on average, reduces our uncertainty over
\(\pbest\) the most. We quantify our uncertainty using the Shannon entropy
\(\,\mathscr{H}(\pbest)\). For a candidate point \(x_i\), let \(\hat{\pi}(c \mid x_i)\)
be the probability that \(x_i\) belongs to class \(c\) under our current beliefs
(as in \cref{eq:classmarginal}, without marginalizing). For each hypothetical label \(c\), we perform a
\emph{virtual} update of all confusion matrix rows according to the update procedure defined in the next section (\cref{sec:update}, \cref{eq:dirichlet-update}). This yields a \emph{hypothetical} distribution \(\pbest^c\). We measure the new
entropy \(\mathscr{H}(\pbest^c)\) and then revert to our original state. Weighting by
\(\hat{\pi}(c \mid x_i)\), the \emph{expected posterior entropy} is
\vspace{-6pt}
\begin{equation}
  \sum_{c=1}^C
    \hat{\pi}\bigl(c \mid x_i\bigr)
    \,\mathscr{H}\bigl(\pbest^c\bigr).
    \vspace{-3pt}
\end{equation}
Hence, the expected information gain (EIG) for point \(x_i\) is
\vspace{-6pt}
\begin{equation}
  \mathrm{EIG}(x_i)
  \;=\;
  \mathscr{H}\bigl(\pbest\bigr)
  \;-\;
  \sum_{c=1}^C
    \hat{\pi}\bigl(c \mid x_i\bigr)\,
    \mathscr{H}\bigl(\pbest^c\bigr).
    \label{eq:eig}
\vspace{-3pt}
\end{equation}
At each iteration, we compute \(\mathrm{EIG}(x_i)\) for all unlabeled points
and query the label for the one with the highest EIG, then perform the
\emph{real} partial update to our Dirichlet parameters as in~\cref{eq:dirichlet-update}.

\subsection{Updating beliefs (\cref{fig:method}, Step 3)}
\label{sec:update}

As true labels become available, we update our confusion matrix estimates to incorporate new information. Consider we have just observed the label for $x_i$ is $y_i =c$. Recall from \cref{sec:ds} that we model each classifier's class prediction for $x_i$ as a random draw according to row $c$ of its confusion matrix, ${\hat c_{k, i} \sim \text{Cat}(M_{k,\, c}})$. We use the fact that our Dirichlet prior for this row is the conjugate prior for this categorical distribution to update the Dirichlet for the next time step as:
\vspace{-10pt}
\begin{align}
  \theta_{k,\,c,\,\hat{c}_{k,i}}
  \leftarrow\
  \theta_{k,\,c,\,\hat{c}_{k,i}} + \eta\
  \label{eq:dirichlet-update}
  \vspace{-5pt}
\end{align}
Where $\eta$ is a learning rate hyperparameter allowing for partial updates. When $\eta=1$ this reduces to the standard Dirchlet-categorical update rule. In practice, we find that partial updates with $\eta<1$ are useful for stability. We use $\eta = 0.01$ by default for all experiments.

\section{Datasets}
\vspace{-5pt}

Models are not re-trained during model selection. Therefore a \textit{model selection benchmark} can be represented as a tuple $( \mathbf{p}, \mathbf{y} )$, where $\mathbf{p} \in \mathbb{R}^{|H|\times|D|\times C}$ is the set of predictions for each model $h \in H$, data point $x_i \in D$, and class $c \in C$, and $\mathbf{y} = \{ y_1, \dots, y_{|D|} \}$ are the ground-truth labels for each data point in $D$. We perform model selection \textit{directly on the test set}, \ie there is no validation/test set split. This matches the pool-based setting of prior work in active testing and active model selection~\cite{kossen2021active,okanovic2024all}.

We curate a suite of 26 diverse model selection benchmarking tasks from 3 different existing benchmarks along with over 3500 pre-trained models. This benchmark suite represents the largest empirical study of active model selection to date.
In this section we describe the models in the candidate pool as well as how they were trained.
In some cases, we train these models ourselves; in others, we source public pre-trained models for which we do not have any information about the training process. We publish our benchmark suite, both the curated set of datasets and the pretrained models, to support future research.

\noindent
\textbf{ModelSelector}~\cite{okanovic2024all} is a benchmark suite focused on pre-trained models sourced from online repositories such as HuggingFace Models and PyTorch hub. We source the prediction files directly from the ModelSelector GitHub repository. 
We curate tasks for which at least 100 test data points are available for easy comparison with other datasets in our benchmarking suite. In total we include ten image and text based classification tasks: 
CIFAR10 (low accuracy models)~\cite{krizhevsky2009learning}, CIFAR10 (high accuracy models), PACS~\cite{li2017deeper}, and seven tasks from the GLUE language classification benchmark~\cite{wang2018glue}. We refer to the vision tasks collectively as MSV (``ModelSelector Vision''). MSV and GLUE involve between 9--114 models per task totaling 851 models.

\noindent
\textbf{WILDS}~\cite{koh2021wilds} is a benchmark of in-the-wild distribution shifts which provides an opportunity to study model selection in the domain generalization setting, where models have been specifically trained to generalize to new test data.
We focus on all classification tasks in WILDS where performing standard model selection using the provided validation sets results in a regret greater than 1\%, \ie where active model selection would be beneficial (experiments included in supplemental).
These tasks are: iWildCam~\cite{beery2019iwildcam}, which involves classifying wildlife in imagery; FMoW~\cite{christie2018functional}, which involves classification of land use in remote sensing imagery; CivilComments~\cite{borkan2019nuanced}, which involves toxicity classification of text data; and Camelyon17~\cite{bandi2018detection}, which involves tumor classification in histopathology data. These tasks range from binary to 182-way classification. 
We train all baseline algorithms from ~\citet{koh2021wilds} using their publicly-available codebase: empirical risk minimization with in-distribution validation (ERM)~\cite{vapnik1999overview}, CORAL~\cite{sun2016deep}, IRM~\cite{arjovsky2019invariant}, and GroupDRO~\cite{hu2018does}. 
We train each method for the default number of epochs used by WILDS, saving a checkpoint every epoch, resulting in between 20 and 240 models per task and 348 models total.

\noindent
\textbf{DomainNet126}~\cite{saito2019semi,peng2019moment}
is an unsupervised domain adaptation benchmark where the task is 126-way classification of objects in real and synthetic imagery. We follow \citet{peng2019moment} and construct 12 adaptation tasks across 4 domains: real imagery, paintings, clipart, and sketches, 
and use the standard UDA training protocol outlined used in prior work~\cite{saito2019semi,peng2019moment,musgrave2021unsupervised}. 
We use the Powerful-Benchmarker codebase~\cite{musgrave2022three} to train 10 popular unsupervised domain algorithms on DomainNet126: ATDOC~\cite{liang2021domain}, BNM~\cite{cui2020towards}, BSP~\cite{chen2019transferability}, CDAN~\cite{long2018conditional}, DANN~\cite{ganin2016domain}, GVB~\cite{cui2020gradually}, IM~\cite{shi2012information}, MCC~\cite{jin2020minimum}, MCD~\cite{saito2018maximum}, and MMD~\cite{long2015learning}. 
We train each method for 40 epochs, saving a checkpoint every 2 epochs.
In total, this gives us 10 algorithms $\times$ 20 checkpoints $=$ 200 models for each transfer task (2400 models overall). 

\vspace{-5pt}
\section{Baselines}
\vspace{-5pt}

We compare our method against five other active model selection methods, ranging from classic approaches to recent state-of-the-art methods. The methods are:

\noindent \textbf{Random sampling} The simplest baseline samples a point uniformly at random each time step and maintains an empirical risk estimate for each model over time. We perform model selection by returning the model with the lowest empirical risk estimate at every time step.

\noindent \textbf{Uncertainty sampling~\cite{dagan1995committee}} We follow \citet{okanovic2024all} to adapt classic committee-based uncertainty sampling techniques from active learning~\cite{dagan1995committee} to the active model selection scenario. At each time step, we greedily sample the point that most models in $H$ disagree on, defined as the entropy of the mean prediction of all models. We perform model selection with empirical risk estimation.

\noindent \textbf{Active Testing~\cite{kossen2021active}} Active Testing aims to obtain label-efficient unbiased estimates of model performance through importance-weighted sampling. To do so, they use a \textit{surrogate model}, which is assumed to be more accurate than any candidate models in $H$, to guide the data sampling process. Points are sampled stochastically in proportion to the estimated loss of the model of interest with respect to the surrogate's predictions. We adapt the framework to the active model selection setting as follows: we instantiate the surrogate model as the same ensemble we use to form our initial consensus estimates. We implement a naive extension of the Active Testing acquisition function that simply sums the acquisition probabilities from all models in the hypothesis set. We perform model selection with empirical risk estimation using unbiased risk estimators~\cite{kossen2021active,farquhar2021statistical}.

\noindent \textbf{VMA~\cite{matsuura2023active}} VMA is an active model selection extension to the Active Testing framework. Their acquisition function is based on minimizing the pairwise variance of the difference between model loss estimates, where the loss estimates are the same importance-weighted estimates as \cite{kossen2021active}. Again we instantiate the surrogate model as the ensemble of all models in $H$ and return the model with the lowest unbiased risk estimate every time step.

\noindent \textbf{ModelSelector~\cite{okanovic2024all}} ModelSelector is currently the state-of-the-art method for active model selection. It utilizes a probability distribution similar to our \pbest, but computes this independently of any per-model performance metrics. Their \pbest is updated according to the following update rule:
$P_{\text{Best}, t+1}(h_k) = \frac{1-\epsilon}{\epsilon} \times P_{\text{Best}, t+1}(h_k)$ if $h_k$ gets the label at time $t$ correct. $\epsilon$ is a learning rate hyperparameter that is set with a self-supervised protocol. We follow this protocol using their codebase to find the optimal $\epsilon$ for any datasets that they do not benchmark in their paper.

\vspace{-1pt}
\section{Experiments}
\label{sec:results}

\noindent \textbf{Experimental settings} All results are reported as the mean over five random seeds. We do not tune the hyperparameters of our method on each dataset; instead we select fixed values $\{ T=0.5, \alpha=0.1, \eta=0.01 \}$ based on a limited set of initial experiments. 

\begin{table}[t]
\centering
\resizebox{\linewidth}{!}{

\begin{tabular}{clrrrrrr}
\toprule

& \multirow{2}{*}{Task} & Random & \multirow{2}{*}{Uncertainty} & Active & \multirow{2}{*}{VMA} & Model & \cellcolor{gray!15}\textbf{CODA} \\
& & Sampling &  & Testing &  & Selector & {\cellcolor{gray!15}\textbf{(Ours)}}\\
\midrule

{\multirow{12}{*}{\rotatebox[origin=c]{90}{DomainNet126}}} & real$\rightarrow$sketch & 147.1 & 197.7 & 269.8 & 119.2 & \textbf{88.8} & \cellcolor{gray!15}\underline{101.2} \\ 
& real$\rightarrow$painting & 143.6 & 167.9 & 118.9 & 139.9 & \underline{92.1} & \cellcolor{gray!15}\textbf{87.2} \\ 
& real$\rightarrow$clipart & 237.7 & 217.6 & \textbf{152.1} & 206.9 & \underline{153.2} & \cellcolor{gray!15}231.7 \\ 
& sketch$\rightarrow$real & 280.4 & 252.7 & \underline{236.4} & 273.4 & 268.4 & \cellcolor{gray!15}\textbf{11.9} \\ 
& sketch$\rightarrow$painting & \underline{156.6} & 181.9 & 237.7 & 157.1 & 173.9 & \cellcolor{gray!15}\textbf{13.0} \\ 
& sketch$\rightarrow$clipart & 228.2 & 224.6 & \underline{162.0} & 227.9 & \textbf{38.1} & \cellcolor{gray!15}432.4 \\ 
& painting$\rightarrow$real & 364.8 & 224.0 & \underline{215.6} & 358.9 & 293.4 & \cellcolor{gray!15}\textbf{2.4} \\ 
& painting$\rightarrow$sketch & \underline{179.3} & 440.7 & 202.5 & 211.5 & 209.8 & \cellcolor{gray!15}\textbf{72.3} \\ 
& painting$\rightarrow$clipart & 222.7 & 296.6 & 251.4 & 271.8 & \underline{73.2} & \cellcolor{gray!15}\textbf{43.1} \\ 
& clipart$\rightarrow$real & 322.1 & 177.1 & 159.1 & 306.4 & \underline{72.7} & \cellcolor{gray!15}\textbf{25.3} \\ 
& clipart$\rightarrow$sketch & \underline{247.2} & 924.8 & 282.6 & 291.3 & 532.7 & \cellcolor{gray!15}\textbf{51.3} \\ 
& clipart$\rightarrow$painting & 147.0 & 162.9 & 222.6 & 167.9 & \underline{131.7} & \cellcolor{gray!15}\textbf{122.2} \\ 
\midrule
{\multirow{4}{*}{\rotatebox[origin=c]{90}{WILDS}}} & iwildcam & \underline{287.0} & 380.4 & 392.1 & 440.6 & 459.0 & \cellcolor{gray!15}\textbf{201.7} \\ 
& camelyon & \underline{175.2} & 311.6 & 206.1 & \textbf{160.1} & 198.3 & \cellcolor{gray!15}288.7 \\ 
& fmow & 189.7 & 191.9 & \underline{153.0} & 189.2 & 211.7 & \cellcolor{gray!15}\textbf{70.0} \\ 
& civilcomments & 140.9 & \textbf{13.3} & 76.7 & \underline{50.1} & 125.3 & \cellcolor{gray!15}318.6 \\ 
\midrule
{\multirow{3}{*}{\rotatebox[origin=c]{90}{MSV}}} & cifar10-low & 410.6 & 629.7 & \underline{399.9} & 476.5 & 567.2 & \cellcolor{gray!15}\textbf{58.7} \\ 
& cifar10-high & 346.2 & 281.7 & 154.9 & 383.7 & \underline{90.9} & \cellcolor{gray!15}\textbf{74.1} \\ 
& pacs & 216.6 & \textbf{6.9} & 101.8 & 116.5 & 97.9 & \cellcolor{gray!15}\underline{57.9} \\ 
\midrule
{\multirow{7}{*}{\rotatebox[origin=c]{90}{GLUE}}} & cola & 368.1 & \textbf{169.5} & 239.8 & 317.8 & \underline{207.8} & \cellcolor{gray!15}2226.7 \\ 
& mnli & 237.4 & \underline{80.8} & 234.2 & 312.8 & 148.5 & \cellcolor{gray!15}\textbf{23.5} \\ 
& qnli & 222.7 & 231.6 & 246.6 & 283.0 & \underline{185.7} & \cellcolor{gray!15}\textbf{120.4} \\ 
& qqp & 136.7 & 388.9 & \underline{127.8} & 169.2 & 186.8 & \cellcolor{gray!15}\textbf{4.8} \\ 
& rte & 375.7 & 390.3 & 424.4 & 674.7 & \textbf{243.6} & \cellcolor{gray!15}\underline{283.8} \\ 
& sst2 & 219.9 & \underline{89.6} & 174.9 & 373.6 & 202.0 & \cellcolor{gray!15}\textbf{51.7} \\ 
& mrpc & 318.2 & 301.1 & 235.2 & 332.3 & \underline{173.1} & \cellcolor{gray!15}\textbf{49.0} \\ 
\bottomrule
\end{tabular}
}

\vspace{-5pt}
\caption{\textbf{Active model selection main results: cumulative regret at step 100.} 
Best method per task in bold, second best underlined (lower is better). CODA is state-of-the-art on 18 out of 26 tasks, often significantly outperforming the next-best method, \eg by $90\times$ on painting$\rightarrow$real. Variances reported in supplemental.} 
\vspace{-10pt}
\label{tab:main-results}
\end{table}

\begin{figure}[t]
    \includegraphics[width=\linewidth]{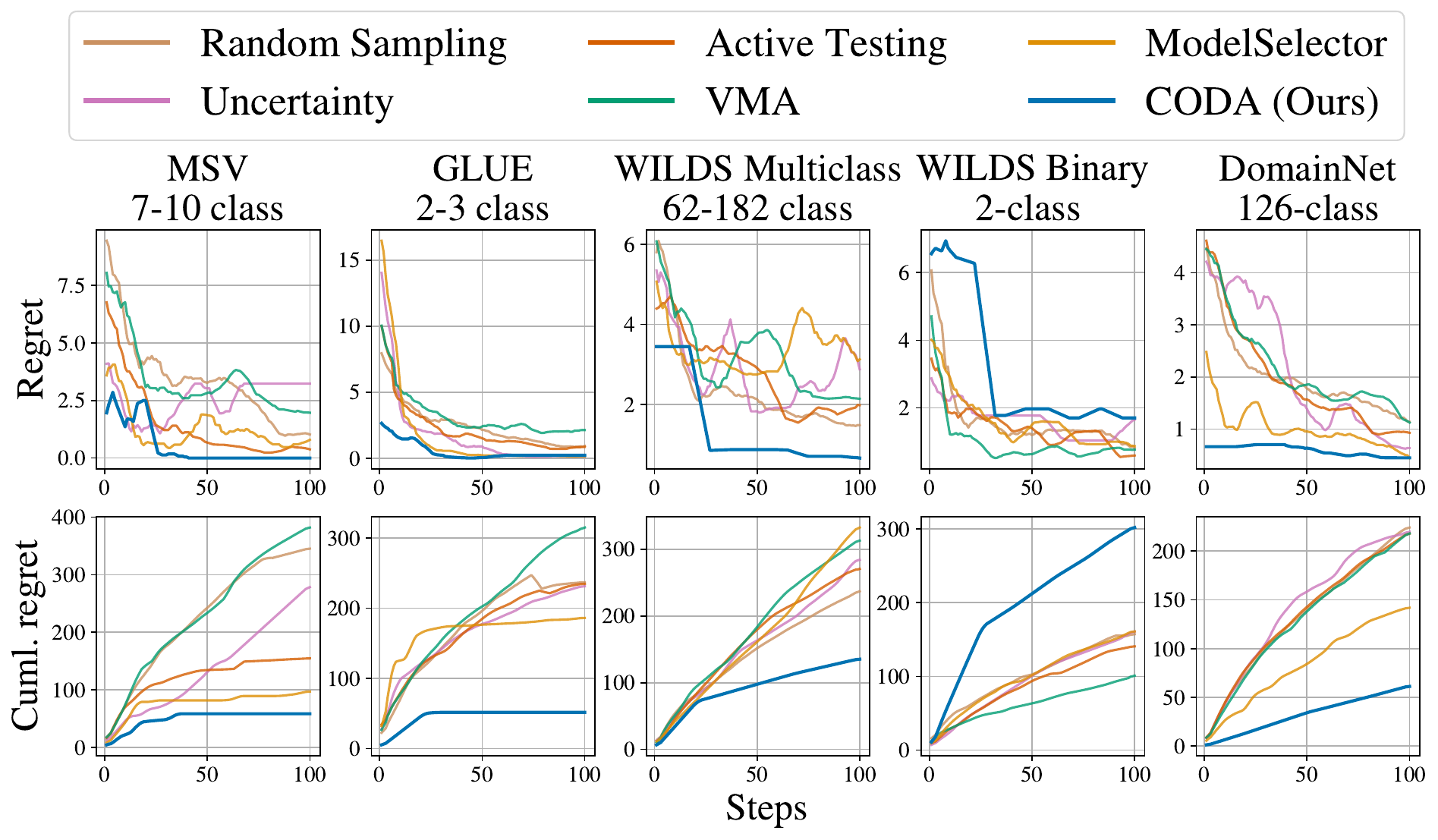}
    \vspace{-23pt}
    \caption{\textbf{Active model selection average results.} We visualize regret (top row) and cumulative regret (bottom row) from time steps 1 to 100, median value across all tasks within benchmarks. Lower is better. CODA is consistently the best performer over time for all settings except the binary classification tasks in WILDS. Full per-task results in supplemental.}
    \label{fig:main-results}
    \vspace{-15pt}
\end{figure}

\noindent \textbf{Active model selection results} Our main results for active model selection are shown in \cref{tab:main-results}. We use accuracy as our loss function and cumulative regret at step 100 as our main point of comparison for all methods, as this provides a summary of overall performance in the few-label regime where active selection is most impactful. We report additional metrics in the supplemental. For a more detailed look at performance within this time window, we visualize the regret and cumulative regret over time in more detail in \cref{fig:main-results} and for all tasks separately in the supplemental.

CODA outperforms all prior work on 18 out of the 26 datasets tested, often significantly, resulting in an over 80\% reduction in regret compared to the next-best method on 5 datasets, greater than 50\% reduction compared to the next-best on 11 datasets, and greater than 25\% reduction from the next-best on 15 datasets. The next-best method is inconsistent across benchmarks. Of the eight datasets where ours is not the best method, uncertainty-based sampling and ModelSelector are best on three each while
Active Testing and VMA are best on one each. Our method performs worst on CoLA and CivilComments, where we underperform random sampling by $6.1\times$ and $2.3\times$, respectively.
We analyze these successes and failures in more detail in the remainder of this section. 

\noindent \textbf{Ablation studies} First we ablate the priors used by our method in \cref{tab:ablation-prior}. 
We see that the consensus-informed priors introduced in \cref{eq:consensus} are a key component of our good performance. Removing them increases regret significantly during the early parts of the label acquisition process. We also see that the diagonal-weighted prior introduced in \cref{eq:diagonal} to regularize the consensus priors performs better than uniform on some, but not all, datasets regardless of the number of classes. Note that in binary classification, the uniform and diagonal settings are equivalent.

In \cref{tab:ablation-q}, we ablate our acquisition function (expected information gain w.r.t. \pbest, \cref{eq:eig}). We compare with random sampling as well as uncertainty-based sampling~\cite{dagan1995committee} which greedily selects the data point with the largest Shannon entropy in the mean prediction of all models. 
We see that expected information gain results in the lowest cumulative regret in most cases. However we also see that uncertainty-based sampling also performs well in combination with the CODA probabilistic framework, sometimes outperforming expected information gain. 

\begin{figure}
    \centering
\includegraphics[width=\linewidth]{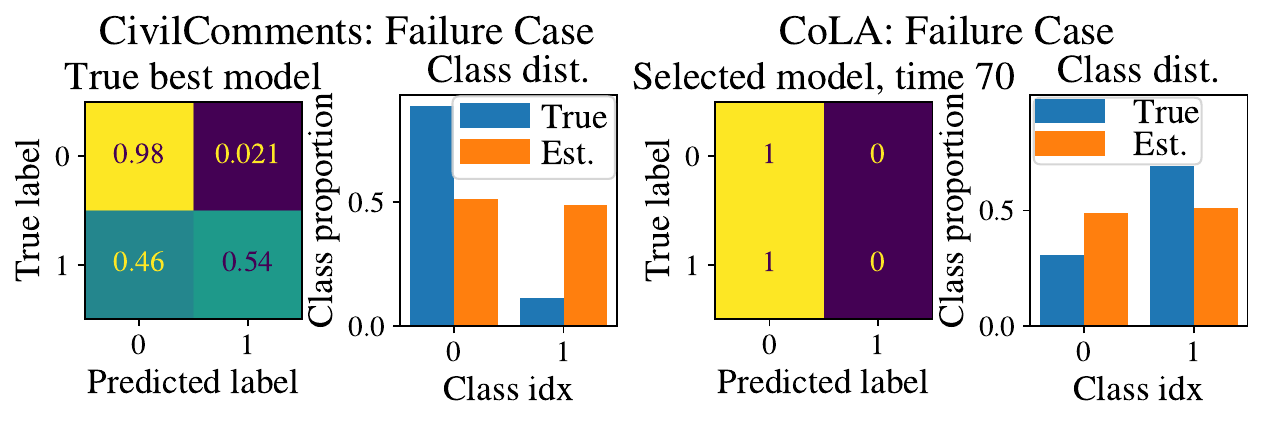}
\vspace{-23pt}
    \caption{\textbf{Failure analysis on CivilComments and CoLA.} 
    CODA may underestimate the performance of very biased classifiers in early steps (CivilComments, left), but overestimate them in later steps (CoLA, right) when there is also data imbalance present (blue bars). More details in \cref{sec:results}.}
    \label{fig:failure}
    \vspace{-10pt}
\end{figure}

\begin{figure}
    \centering
    \includegraphics[width=\linewidth]{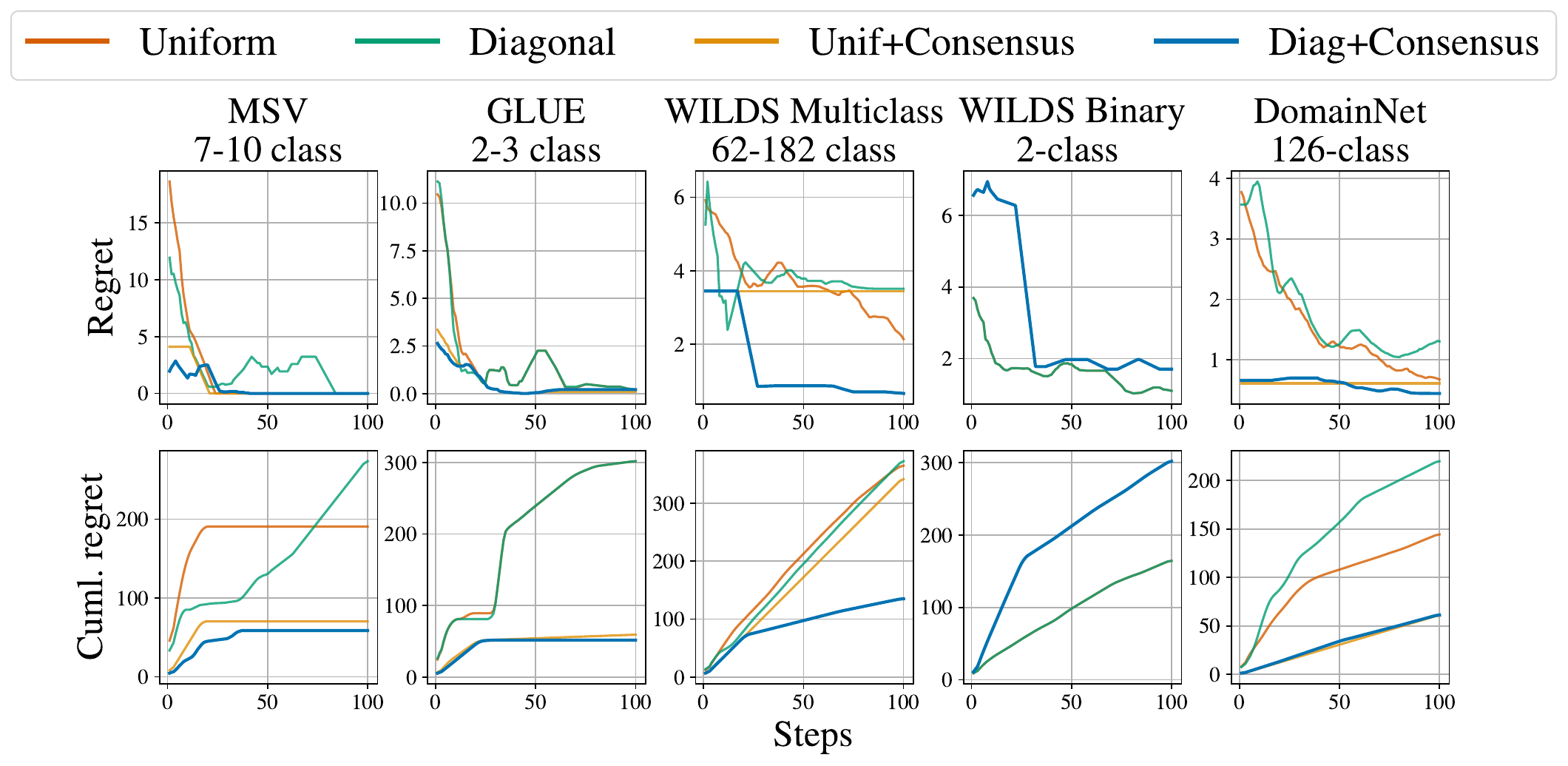}
    \vspace{-20pt}
    \caption{\textbf{Ablation of CODA prior design.} We compare a uniform prior on the confusion matrices (top row) with diagonal upweighting (``Diag.'' column; \cref{eq:diagonal}) and consensus prior (``Cons.'' column; \cref{eq:consensus}) we introduce. 
    For binary classification tasks, uniform and diagonal weighting are equivalent.
    In most cases, both the consensus prior and diagonal upweighting provide benefits, and are complementary.}
    \label{tab:ablation-prior}
    \vspace{-5pt}
\end{figure}

\begin{figure}
    \centering
    \includegraphics[width=\linewidth]{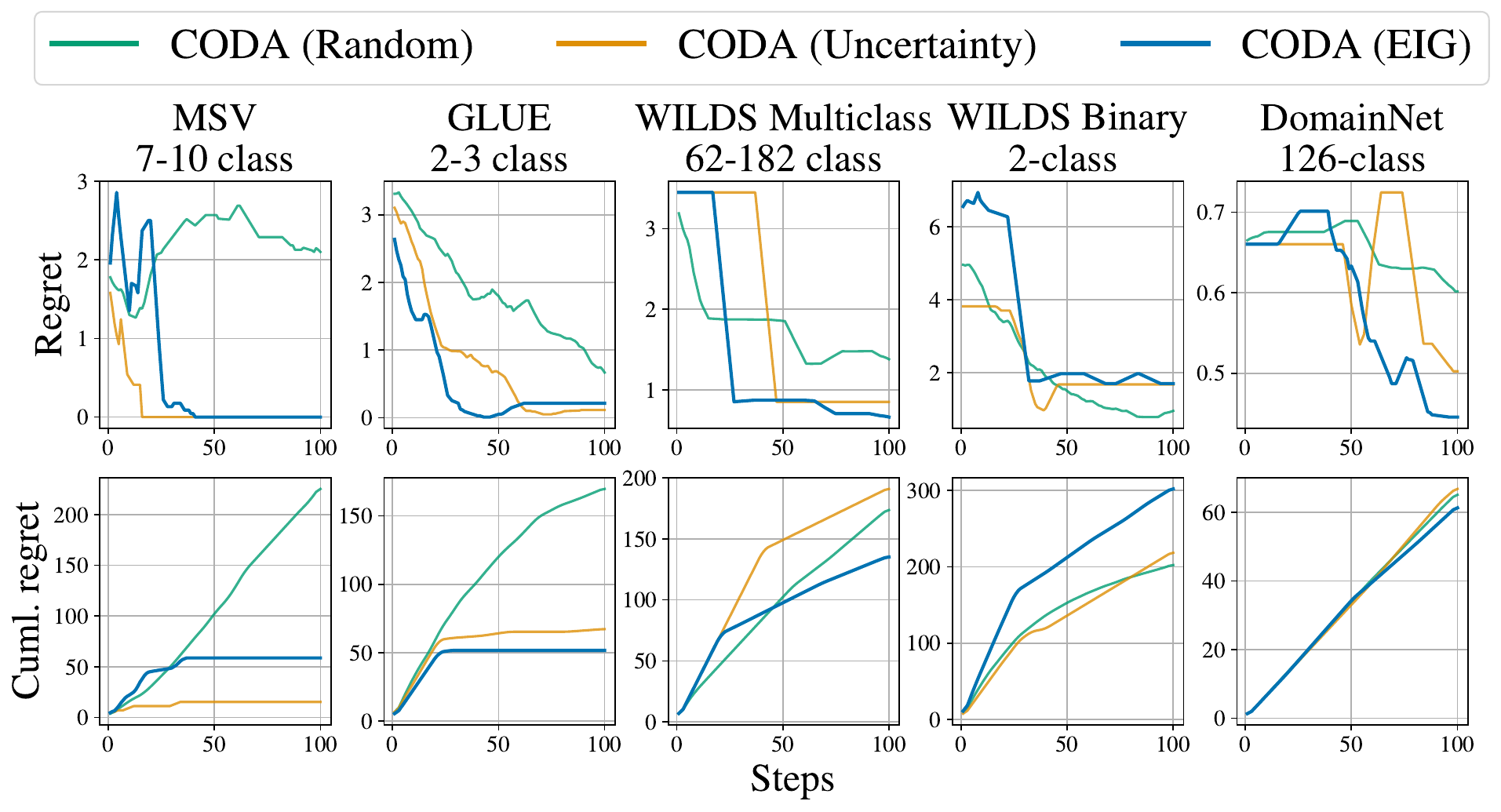}
    \vspace{-20pt}
    \caption{\textbf{Ablation of CODA acquisition function.} We use the CODA probabilistic framework and compare different data point acquisition functions: random sampling, uncertainty-based sampling, and expected information gain (EIG, the default).  We see that EIG typically improves upon other sampling approaches, however uncertainty-based sampling is also a strong acquisition function in combination with the rest of the CODA framework.}
    \label{tab:ablation-q}
    \vspace{-10pt}
\end{figure}

\noindent \textbf{Limitations and failure analysis} We investigate CODA's poor performance on CivilComments and CoLA in \cref{fig:failure}. We find that poor performance in these cases is caused by a combination of data imbalance and model bias. CODA uses the predicted class marginal in its updates (\cref{eq:eig}), but regularizes this to counteract overconfident predictions (\cref{eq:diagonal}). In CivilComments this causes us to upweight the contribution of minority class predictions early on, requiring a moderate amount of samples to identify the imbalance. 
In addition, the best model on CivilComments is extremely biased, with 98\% accuracy on the majority class but only 54\% accuracy on the minority class, exacerbating the problem when selecting based on micro-accuracy.
Similarly, in CoLA, CODA selects a very biased model---one that only predicts the negative class---\textit{later} in the model selection process (step 70) again because the dataset is estimated to be more balanced than it actually is.
We note that these failures are exceptional cases, as CODA performs well on other datasets with significant imbalance (\eg iWildCam). 
This points to interesting directions for future work to address data imbalance, model bias, and use-case specific metrics.

\noindent \textbf{Additional unsupervised results} Though not the main focus of this paper, our method can also be used to perform \textit{unsupervised} model selection by using only the consensus-informed priors to compute \pbest. We show that this is remarkably effective, matching or outperforming the previous state-of-the-art in unsupervised model selection in 20 of 26 benchmarks, in the supplemental material.

\vspace{-5pt}
\section{Conclusion and future work}
\vspace{-5pt}

We have introduced a new method for active model selection that can identify the best model in a pool of candidates with significantly fewer human labels than prior work. The ability to do so has implications for users of pre-trained machine learning models, who can use CODA to efficiently select the best model for their dataset, as well as for researchers in fields such as domain adaptation who face model selection challenges during model development.

Our work opens up several exciting areas for future research. Within the active model selection framework there are several interesting directions: (1) How to better construct and utilize informative priors, whether from unsupervised procedures or by incorporating human-provided domain knowledge; (2) Extending the active model selection framework beyond accuracy and classification to support more tasks and target metrics; (3) Exploring more sophisticated probabilistic models that better capture predictive quantities such as model confidence.

More broadly, active model selection can be seen as one facet of a larger research goal of how to best utilize human effort in the development and deployment of machine learning systems. We show that active model selection is a particularly effective use of annotation effort, but there are many things that \textit{could} be done with collected labels, either in serial or in parallel. For example, an interesting research direction is how to perform active learning and/or active testing concurrently with active model selection, and how to efficiently allocate effort to the different tasks. We hope our work can provide a strong starting point for investigating these questions.

\section*{Acknowledgments}

We would like to thank Julia Chae, Mark Hamilton, Timm Haucke, Michael Hobley, Neha Hulkund, Rupa Kurinchi-Vendhan, Evan Shelhamer, and Edward Vendrow for feedback on early drafts, and Serge Belongie, Emma Pierson, Manish Raghavan, Shuvom Sadhuka, and Divya Shanmugam for helpful discussions. This work was supported in part by NSF awards \#2313998, \#2330423, and \#2329927, NSERC award \#585136, and MIT J-WAFS seed grant \#2040131. 

{
    \small
    \bibliographystyle{ieeenat_fullname}
    \bibliography{main}
}


\clearpage

\section*{Supplemental Material}

\section{Additional results}

\subsection{Alternate metrics}

Here we report tabular results for several alternate metrics, providing additional points of comparison to supplement our main results of cumulative regret at step 100 (\cref{tab:main-results}).

\subsubsection{Variance between seeds}

\begin{table}[h]
\centering
\resizebox{\linewidth}{!}{

\begin{tabular}{clrrrrrr}
\toprule

& \multirow{2}{*}{Task} & Random & \multirow{2}{*}{Uncertainty} & Active & \multirow{2}{*}{VMA} & Model & \cellcolor{gray!15}\textbf{CODA} \\
& & Sampling &  & Testing &  & Selector & {\cellcolor{gray!15}\textbf{(Ours)}}\\
\midrule

{\multirow{12}{*}{\rotatebox[origin=c]{90}{DomainNet126}}} & real$\rightarrow$sketch & 147.1{\small $\pm$ 38.5} & 197.7{\small $\pm$ 13.0} & 269.8{\small $\pm$ 119.8} & 119.2{\small $\pm$ 51.9} & \textbf{88.8{\small $\pm$ 9.8}} & \cellcolor{gray!15}\underline{101.2{\small $\pm$ 0.0}} \\ 
& real$\rightarrow$painting & 143.6{\small $\pm$ 37.1} & 167.9{\small $\pm$ 6.8} & 118.9{\small $\pm$ 68.8} & 139.9{\small $\pm$ 53.3} & \underline{92.1{\small $\pm$ 1.6}} & \cellcolor{gray!15}\textbf{87.2{\small $\pm$ 2.8}} \\ 
& real$\rightarrow$clipart & 237.7{\small $\pm$ 88.7} & 217.6{\small $\pm$ 23.7} & \textbf{152.1{\small $\pm$ 40.0}} & 206.9{\small $\pm$ 100.7} & \underline{153.2{\small $\pm$ 9.5}} & \cellcolor{gray!15}231.7{\small $\pm$ 0.0} \\ 
& sketch$\rightarrow$real & 280.4{\small $\pm$ 164.4} & 252.7{\small $\pm$ 11.8} & \underline{236.4{\small $\pm$ 160.5}} & 273.4{\small $\pm$ 132.7} & 268.4{\small $\pm$ 7.5} & \cellcolor{gray!15}\textbf{11.9{\small $\pm$ 0.0}} \\ 
& sketch$\rightarrow$painting & \underline{156.6{\small $\pm$ 72.6}} & 181.9{\small $\pm$ 7.3} & 237.7{\small $\pm$ 151.7} & 157.1{\small $\pm$ 60.9} & 173.9{\small $\pm$ 7.1} & \cellcolor{gray!15}\textbf{13.0{\small $\pm$ 0.0}} \\ 
& sketch$\rightarrow$clipart & 228.2{\small $\pm$ 74.5} & 224.6{\small $\pm$ 43.2} & \underline{162.0{\small $\pm$ 83.6}} & 227.9{\small $\pm$ 85.6} & \textbf{38.1{\small $\pm$ 9.2}} & \cellcolor{gray!15}432.4{\small $\pm$ 1.9} \\ 
& painting$\rightarrow$real & 364.8{\small $\pm$ 164.4} & 224.0{\small $\pm$ 7.2} & \underline{215.6{\small $\pm$ 94.4}} & 358.9{\small $\pm$ 193.6} & 293.4{\small $\pm$ 20.3} & \cellcolor{gray!15}\textbf{2.4{\small $\pm$ 0.0}} \\ 
& painting$\rightarrow$sketch & \underline{179.3{\small $\pm$ 53.5}} & 440.7{\small $\pm$ 32.9} & 202.5{\small $\pm$ 79.7} & 211.5{\small $\pm$ 73.3} & 209.8{\small $\pm$ 5.2} & \cellcolor{gray!15}\textbf{72.3{\small $\pm$ 0.2}} \\ 
& painting$\rightarrow$clipart & 222.7{\small $\pm$ 144.5} & 296.6{\small $\pm$ 6.1} & 251.4{\small $\pm$ 111.8} & 271.8{\small $\pm$ 116.9} & \underline{73.2{\small $\pm$ 3.2}} & \cellcolor{gray!15}\textbf{43.1{\small $\pm$ 0.1}} \\ 
& clipart$\rightarrow$real & 322.1{\small $\pm$ 127.8} & 177.1{\small $\pm$ 34.6} & 159.1{\small $\pm$ 55.6} & 306.4{\small $\pm$ 181.7} & \underline{72.7{\small $\pm$ 15.3}} & \cellcolor{gray!15}\textbf{25.3{\small $\pm$ 0.0}} \\ 
& clipart$\rightarrow$sketch & \underline{247.2{\small $\pm$ 143.0}} & 924.8{\small $\pm$ 22.7} & 282.6{\small $\pm$ 186.5} & 291.3{\small $\pm$ 163.9} & 532.7{\small $\pm$ 76.1} & \cellcolor{gray!15}\textbf{51.3{\small $\pm$ 0.0}} \\ 
& clipart$\rightarrow$painting & 147.0{\small $\pm$ 46.2} & 162.9{\small $\pm$ 13.3} & 222.6{\small $\pm$ 100.5} & 167.9{\small $\pm$ 100.2} & \underline{131.7{\small $\pm$ 2.5}} & \cellcolor{gray!15}\textbf{122.2{\small $\pm$ 0.1}} \\ 
\midrule
{\multirow{4}{*}{\rotatebox[origin=c]{90}{WILDS}}} & iwildcam & \underline{287.0{\small $\pm$ 65.1}} & 380.4{\small $\pm$ 7.1} & 392.1{\small $\pm$ 194.1} & 440.6{\small $\pm$ 103.6} & 459.0{\small $\pm$ 56.0} & \cellcolor{gray!15}\textbf{201.7{\small $\pm$ 0.0}} \\ 
& camelyon & \underline{175.2{\small $\pm$ 81.7}} & 311.6{\small $\pm$ 11.9} & 206.1{\small $\pm$ 121.1} & \textbf{160.1{\small $\pm$ 76.8}} & 198.3{\small $\pm$ 90.6} & \cellcolor{gray!15}288.7{\small $\pm$ 0.0} \\ 
& fmow & 189.7{\small $\pm$ 66.8} & 191.9{\small $\pm$ 8.4} & \underline{153.0{\small $\pm$ 38.7}} & 189.2{\small $\pm$ 62.6} & 211.7{\small $\pm$ 28.1} & \cellcolor{gray!15}\textbf{70.0{\small $\pm$ 0.2}} \\ 
& civilcomments & 140.9{\small $\pm$ 75.5} & \textbf{13.3{\small $\pm$ 2.6}} & 76.7{\small $\pm$ 119.0} & \underline{50.1{\small $\pm$ 29.3}} & 125.3{\small $\pm$ 59.6} & \cellcolor{gray!15}318.6{\small $\pm$ 0.0} \\ 
\midrule
{\multirow{3}{*}{\rotatebox[origin=c]{90}{MSV}}} & cifar10-low & 410.6{\small $\pm$ 231.1} & 629.7{\small $\pm$ 5.9} & \underline{399.9{\small $\pm$ 207.5}} & 476.5{\small $\pm$ 114.4} & 567.2{\small $\pm$ 23.9} & \cellcolor{gray!15}\textbf{58.7{\small $\pm$ 0.0}} \\ 
& cifar10-high & 346.2{\small $\pm$ 147.4} & 281.7{\small $\pm$ 21.3} & 154.9{\small $\pm$ 68.9} & 383.7{\small $\pm$ 141.9} & \underline{90.9{\small $\pm$ 12.3}} & \cellcolor{gray!15}\textbf{74.1{\small $\pm$ 0.0}} \\ 
& pacs & 216.6{\small $\pm$ 86.8} & \textbf{6.9{\small $\pm$ 5.4}} & 101.8{\small $\pm$ 55.9} & 116.5{\small $\pm$ 57.5} & 97.9{\small $\pm$ 6.0} & \cellcolor{gray!15}\underline{57.9{\small $\pm$ 0.0}} \\ 
\midrule
{\multirow{7}{*}{\rotatebox[origin=c]{90}{GLUE}}} & cola & 368.1{\small $\pm$ 80.6} & \textbf{169.5{\small $\pm$ 27.1}} & 239.8{\small $\pm$ 162.4} & 317.8{\small $\pm$ 191.8} & \underline{207.8{\small $\pm$ 61.1}} & \cellcolor{gray!15}2226.7{\small $\pm$ 0.0} \\ 
& mnli & 237.4{\small $\pm$ 132.2} & \underline{80.8{\small $\pm$ 27.9}} & 234.2{\small $\pm$ 105.2} & 312.8{\small $\pm$ 111.5} & 148.5{\small $\pm$ 114.4} & \cellcolor{gray!15}\textbf{23.5{\small $\pm$ 0.0}} \\ 
& qnli & 222.7{\small $\pm$ 68.5} & 231.6{\small $\pm$ 76.2} & 246.6{\small $\pm$ 112.4} & 283.0{\small $\pm$ 78.8} & \underline{185.7{\small $\pm$ 89.6}} & \cellcolor{gray!15}\textbf{120.4{\small $\pm$ 0.0}} \\ 
& qqp & 136.7{\small $\pm$ 20.9} & 388.9{\small $\pm$ 492.8} & \underline{127.8{\small $\pm$ 16.5}} & 169.2{\small $\pm$ 114.1} & 186.8{\small $\pm$ 137.5} & \cellcolor{gray!15}\textbf{4.8{\small $\pm$ 0.0}} \\ 
& rte & 375.7{\small $\pm$ 184.6} & 390.3{\small $\pm$ 68.4} & 424.4{\small $\pm$ 244.4} & 674.7{\small $\pm$ 186.7} & \textbf{243.6{\small $\pm$ 73.2}} & \cellcolor{gray!15}\underline{283.8{\small $\pm$ 0.0}} \\ 
& sst2 & 219.9{\small $\pm$ 108.1} & \underline{89.6{\small $\pm$ 39.6}} & 174.9{\small $\pm$ 54.3} & 373.6{\small $\pm$ 128.6} & 202.0{\small $\pm$ 131.4} & \cellcolor{gray!15}\textbf{51.7{\small $\pm$ 0.0}} \\ 
& mrpc & 318.2{\small $\pm$ 100.3} & 301.1{\small $\pm$ 64.4} & 235.2{\small $\pm$ 36.0} & 332.3{\small $\pm$ 156.0} & \underline{173.1{\small $\pm$ 47.7}} & \cellcolor{gray!15}\textbf{49.0{\small $\pm$ 0.0}} \\ 
\bottomrule
\end{tabular}
}

\caption{\textbf{Main results with variance between seeds: Cumulative regret at step 100.} Same as \cref{tab:main-results} but $\pm$ one standard deviation between runs with different random seeds (5 random seeds used). Standard deviation of 0.0 (as in many CODA runs) indicates method is not stochastic for that task, \eg because of asymmetric priors resulting in deterministic selection.}

\end{table}

\subsubsection{Instantaneous regret}

We report the average instantaneous regret of each method at steps 50 and 100 in \cref{tab:instant50} and \cref{tab:instant100}, respectively. These results are more influenced by stochasticity than cumulative regret, \ie instantaneous regret may fluctuate widely between steps (see \cref{fig:all-results}).

\begin{table}[h]
\centering
\resizebox{\linewidth}{!}{

\begin{tabular}{clrrrrrr}
\toprule

& \multirow{2}{*}{Task} & Random & \multirow{2}{*}{Uncertainty} & Active & \multirow{2}{*}{VMA} & Model & \cellcolor{gray!15}\textbf{CODA} \\
& & Sampling &  & Testing &  & Selector & {\cellcolor{gray!15}\textbf{(Ours)}}\\
\midrule

{\multirow{12}{*}{\rotatebox[origin=c]{90}{DomainNet126}}} & real$\rightarrow$sketch & 1.3{\small $\pm$ 0.6} & 2.0{\small $\pm$ 0.2} & \underline{0.8{\small $\pm$ 0.6}} & 1.3{\small $\pm$ 1.2} & \textbf{0.6{\small $\pm$ 0.0}} & \cellcolor{gray!15}1.0{\small $\pm$ 0.0} \\ 
& real$\rightarrow$painting & 1.2{\small $\pm$ 0.4} & \underline{1.2{\small $\pm$ 0.5}} & 1.4{\small $\pm$ 1.4} & 1.5{\small $\pm$ 1.0} & \textbf{0.9{\small $\pm$ 0.3}} & \cellcolor{gray!15}\underline{1.2{\small $\pm$ 0.0}} \\ 
& real$\rightarrow$clipart & 1.6{\small $\pm$ 1.8} & \underline{0.5{\small $\pm$ 0.6}} & 1.0{\small $\pm$ 0.9} & 0.8{\small $\pm$ 0.8} & \textbf{0.2{\small $\pm$ 0.1}} & \cellcolor{gray!15}2.3{\small $\pm$ 0.0} \\ 
& sketch$\rightarrow$real & 2.0{\small $\pm$ 3.0} & \underline{0.4{\small $\pm$ 0.2}} & 3.0{\small $\pm$ 3.0} & 2.6{\small $\pm$ 3.1} & 5.4{\small $\pm$ 0.0} & \cellcolor{gray!15}\textbf{0.1{\small $\pm$ 0.0}} \\ 
& sketch$\rightarrow$painting & 1.1{\small $\pm$ 1.3} & 0.9{\small $\pm$ 1.1} & 2.2{\small $\pm$ 2.3} & \underline{0.9{\small $\pm$ 0.6}} & 5.5{\small $\pm$ 0.0} & \cellcolor{gray!15}\textbf{0.1{\small $\pm$ 0.0}} \\ 
& sketch$\rightarrow$clipart & 1.8{\small $\pm$ 1.4} & \underline{0.2{\small $\pm$ 0.2}} & 1.6{\small $\pm$ 1.6} & 2.1{\small $\pm$ 0.8} & \textbf{0.0{\small $\pm$ 0.0}} & \cellcolor{gray!15}4.8{\small $\pm$ 0.0} \\ 
& painting$\rightarrow$real & 4.1{\small $\pm$ 2.9} & 2.1{\small $\pm$ 0.7} & \underline{1.3{\small $\pm$ 0.9}} & 2.3{\small $\pm$ 1.1} & 4.6{\small $\pm$ 4.6} & \cellcolor{gray!15}\textbf{0.0{\small $\pm$ 0.0}} \\ 
& painting$\rightarrow$sketch & \underline{1.5{\small $\pm$ 1.8}} & 4.3{\small $\pm$ 0.0} & 2.3{\small $\pm$ 1.5} & 1.9{\small $\pm$ 1.5} & 1.8{\small $\pm$ 0.7} & \cellcolor{gray!15}\textbf{0.9{\small $\pm$ 0.0}} \\ 
& painting$\rightarrow$clipart & 1.9{\small $\pm$ 2.3} & 2.8{\small $\pm$ 0.0} & 3.2{\small $\pm$ 1.7} & 2.8{\small $\pm$ 1.5} & \underline{0.7{\small $\pm$ 0.0}} & \cellcolor{gray!15}\textbf{0.4{\small $\pm$ 0.0}} \\ 
& clipart$\rightarrow$real & 2.6{\small $\pm$ 1.7} & 0.4{\small $\pm$ 0.0} & \underline{0.2{\small $\pm$ 0.4}} & 2.8{\small $\pm$ 1.8} & \textbf{0.1{\small $\pm$ 0.0}} & \cellcolor{gray!15}0.3{\small $\pm$ 0.0} \\ 
& clipart$\rightarrow$sketch & 3.0{\small $\pm$ 2.3} & 8.7{\small $\pm$ 2.9} & \underline{0.8{\small $\pm$ 0.7}} & 2.6{\small $\pm$ 2.0} & 5.4{\small $\pm$ 0.0} & \cellcolor{gray!15}\textbf{0.4{\small $\pm$ 0.0}} \\ 
& clipart$\rightarrow$painting & 1.6{\small $\pm$ 0.9} & \underline{1.1{\small $\pm$ 0.0}} & 2.6{\small $\pm$ 1.9} & 1.3{\small $\pm$ 1.5} & \textbf{1.0{\small $\pm$ 0.1}} & \cellcolor{gray!15}1.2{\small $\pm$ 0.0} \\ 
\midrule
{\multirow{4}{*}{\rotatebox[origin=c]{90}{WILDS}}} & iwildcam & \underline{2.1{\small $\pm$ 0.9}} & 3.0{\small $\pm$ 0.0} & 4.5{\small $\pm$ 2.3} & 6.1{\small $\pm$ 1.9} & 4.6{\small $\pm$ 1.3} & \cellcolor{gray!15}\textbf{0.9{\small $\pm$ 0.0}} \\ 
& camelyon & \textbf{0.5{\small $\pm$ 0.2}} & 3.5{\small $\pm$ 0.0} & 2.1{\small $\pm$ 1.7} & 0.9{\small $\pm$ 1.0} & 1.5{\small $\pm$ 0.8} & \cellcolor{gray!15}\underline{0.6{\small $\pm$ 0.0}} \\ 
& fmow & 1.4{\small $\pm$ 1.2} & \textbf{0.6{\small $\pm$ 0.0}} & 1.2{\small $\pm$ 0.7} & 1.6{\small $\pm$ 1.0} & 1.0{\small $\pm$ 0.0} & \cellcolor{gray!15}\underline{0.8{\small $\pm$ 0.0}} \\ 
& civilcomments & 0.9{\small $\pm$ 1.1} & \textbf{0.0{\small $\pm$ 0.0}} & 0.8{\small $\pm$ 1.5} & \underline{0.3{\small $\pm$ 0.3}} & 1.6{\small $\pm$ 2.7} & \cellcolor{gray!15}3.3{\small $\pm$ 0.0} \\ 
\midrule
{\multirow{3}{*}{\rotatebox[origin=c]{90}{MSV}}} & cifar10-low & 5.5{\small $\pm$ 4.1} & 7.7{\small $\pm$ 0.0} & 5.3{\small $\pm$ 3.4} & \underline{4.7{\small $\pm$ 2.6}} & 7.5{\small $\pm$ 0.4} & \cellcolor{gray!15}\textbf{0.0{\small $\pm$ 0.0}} \\ 
& cifar10-high & 3.1{\small $\pm$ 2.3} & 3.2{\small $\pm$ 0.0} & 0.2{\small $\pm$ 0.2} & 2.6{\small $\pm$ 1.7} & \textbf{0.0{\small $\pm$ 0.0}} & \cellcolor{gray!15}\textbf{0.0{\small $\pm$ 0.0}} \\ 
& pacs & 2.2{\small $\pm$ 1.9} & \textbf{0.0{\small $\pm$ 0.0}} & 0.6{\small $\pm$ 0.8} & 0.8{\small $\pm$ 0.7} & 1.4{\small $\pm$ 0.0} & \cellcolor{gray!15}\textbf{0.0{\small $\pm$ 0.0}} \\ 
\midrule
{\multirow{7}{*}{\rotatebox[origin=c]{90}{GLUE}}} & cola & 3.6{\small $\pm$ 1.3} & 2.0{\small $\pm$ 0.9} & 2.3{\small $\pm$ 1.9} & 2.5{\small $\pm$ 1.9} & \underline{1.4{\small $\pm$ 1.0}} & \cellcolor{gray!15}\textbf{0.4{\small $\pm$ 0.0}} \\ 
& mnli & 2.6{\small $\pm$ 3.0} & \textbf{0.1{\small $\pm$ 0.0}} & 1.0{\small $\pm$ 1.7} & 2.6{\small $\pm$ 1.6} & \textbf{0.1{\small $\pm$ 0.1}} & \cellcolor{gray!15}\textbf{0.1{\small $\pm$ 0.0}} \\ 
& qnli & 1.3{\small $\pm$ 1.3} & 0.9{\small $\pm$ 1.2} & 1.0{\small $\pm$ 1.4} & 1.7{\small $\pm$ 1.5} & \underline{0.1{\small $\pm$ 0.2}} & \cellcolor{gray!15}\textbf{0.0{\small $\pm$ 0.0}} \\ 
& qqp & 1.4{\small $\pm$ 0.2} & 1.3{\small $\pm$ 2.8} & 1.1{\small $\pm$ 0.1} & 1.0{\small $\pm$ 0.6} & \underline{0.3{\small $\pm$ 0.4}} & \cellcolor{gray!15}\textbf{0.0{\small $\pm$ 0.0}} \\ 
& rte & 3.2{\small $\pm$ 4.4} & \textbf{0.0{\small $\pm$ 0.0}} & 3.7{\small $\pm$ 3.6} & 5.9{\small $\pm$ 3.9} & \textbf{0.0{\small $\pm$ 0.0}} & \cellcolor{gray!15}\textbf{0.0{\small $\pm$ 0.0}} \\ 
& sst2 & 1.9{\small $\pm$ 2.5} & 0.2{\small $\pm$ 0.0} & 0.1{\small $\pm$ 0.1} & 3.1{\small $\pm$ 3.5} & \textbf{0.0{\small $\pm$ 0.0}} & \cellcolor{gray!15}\textbf{0.0{\small $\pm$ 0.0}} \\ 
& mrpc & 2.0{\small $\pm$ 2.4} & 1.6{\small $\pm$ 1.9} & 1.6{\small $\pm$ 1.2} & 2.8{\small $\pm$ 2.4} & \underline{0.8{\small $\pm$ 0.5}} & \cellcolor{gray!15}\textbf{0.5{\small $\pm$ 0.0}} \\ 
\bottomrule
\end{tabular}

}
\caption{\textbf{Average instantaneous regret at step 50.} Mean and standard deviation reported over 5 random seeds.}
\label{tab:instant50}
\end{table}

\begin{table}[h]
\centering
\resizebox{\linewidth}{!}{

\begin{tabular}{clrrrrrr}
\toprule

& \multirow{2}{*}{Task} & Random & \multirow{2}{*}{Uncertainty} & Active & \multirow{2}{*}{VMA} & Model & \cellcolor{gray!15}\textbf{CODA} \\
& & Sampling &  & Testing &  & Selector & {\cellcolor{gray!15}\textbf{(Ours)}}\\
\midrule

{\multirow{12}{*}{\rotatebox[origin=c]{90}{DomainNet126}}} & real$\rightarrow$sketch & 1.2{\small $\pm$ 1.1} & \textbf{0.0{\small $\pm$ 0.0}} & 2.2{\small $\pm$ 2.2} & 1.0{\small $\pm$ 0.6} & \underline{0.8{\small $\pm$ 0.5}} & \cellcolor{gray!15}1.0{\small $\pm$ 0.0} \\ 
& real$\rightarrow$painting & 1.1{\small $\pm$ 0.6} & \underline{0.6{\small $\pm$ 0.0}} & 0.8{\small $\pm$ 0.2} & 0.9{\small $\pm$ 0.6} & 1.3{\small $\pm$ 0.3} & \cellcolor{gray!15}\textbf{0.2{\small $\pm$ 0.0}} \\ 
& real$\rightarrow$clipart & \underline{0.6{\small $\pm$ 0.4}} & 1.0{\small $\pm$ 0.0} & 0.6{\small $\pm$ 0.3} & 1.5{\small $\pm$ 2.0} & \textbf{0.2{\small $\pm$ 0.0}} & \cellcolor{gray!15}2.3{\small $\pm$ 0.0} \\ 
& sketch$\rightarrow$real & 1.7{\small $\pm$ 2.8} & 0.5{\small $\pm$ 0.0} & 0.7{\small $\pm$ 1.1} & 0.2{\small $\pm$ 0.1} & \textbf{0.1{\small $\pm$ 0.0}} & \cellcolor{gray!15}\textbf{0.1{\small $\pm$ 0.0}} \\ 
& sketch$\rightarrow$painting & 0.7{\small $\pm$ 0.7} & \textbf{0.0{\small $\pm$ 0.0}} & 0.8{\small $\pm$ 0.8} & 0.8{\small $\pm$ 0.9} & \underline{0.1{\small $\pm$ 0.1}} & \cellcolor{gray!15}\underline{0.1{\small $\pm$ 0.0}} \\ 
& sketch$\rightarrow$clipart & 0.9{\small $\pm$ 0.8} & \underline{0.2{\small $\pm$ 0.3}} & 0.9{\small $\pm$ 0.5} & 2.1{\small $\pm$ 2.0} & \textbf{0.0{\small $\pm$ 0.0}} & \cellcolor{gray!15}0.5{\small $\pm$ 0.0} \\ 
& painting$\rightarrow$real & 3.3{\small $\pm$ 2.4} & 1.2{\small $\pm$ 0.0} & \underline{1.1{\small $\pm$ 0.9}} & 1.4{\small $\pm$ 0.8} & 1.2{\small $\pm$ 0.0} & \cellcolor{gray!15}\textbf{0.0{\small $\pm$ 0.0}} \\ 
& painting$\rightarrow$sketch & \underline{0.6{\small $\pm$ 0.5}} & 0.7{\small $\pm$ 0.0} & 0.7{\small $\pm$ 0.7} & 1.3{\small $\pm$ 1.9} & 1.7{\small $\pm$ 0.0} & \cellcolor{gray!15}\textbf{0.5{\small $\pm$ 0.0}} \\ 
& painting$\rightarrow$clipart & 2.0{\small $\pm$ 1.7} & 0.5{\small $\pm$ 0.2} & 2.3{\small $\pm$ 1.7} & 1.7{\small $\pm$ 1.6} & \textbf{0.3{\small $\pm$ 0.0}} & \cellcolor{gray!15}\underline{0.4{\small $\pm$ 0.0}} \\ 
& clipart$\rightarrow$real & 1.4{\small $\pm$ 1.6} & 0.6{\small $\pm$ 0.0} & 0.7{\small $\pm$ 0.8} & 0.7{\small $\pm$ 0.6} & \textbf{0.1{\small $\pm$ 0.0}} & \cellcolor{gray!15}\underline{0.3{\small $\pm$ 0.0}} \\ 
& clipart$\rightarrow$sketch & 1.6{\small $\pm$ 2.0} & 5.3{\small $\pm$ 2.1} & 1.0{\small $\pm$ 0.7} & \textbf{0.6{\small $\pm$ 0.4}} & \textbf{0.6{\small $\pm$ 0.0}} & \cellcolor{gray!15}\textbf{0.6{\small $\pm$ 0.0}} \\ 
& clipart$\rightarrow$painting & \textbf{0.6{\small $\pm$ 0.5}} & 1.1{\small $\pm$ 0.0} & 1.4{\small $\pm$ 0.7} & 1.1{\small $\pm$ 0.3} & \underline{1.0{\small $\pm$ 0.1}} & \cellcolor{gray!15}1.1{\small $\pm$ 0.0} \\ 
\midrule
{\multirow{4}{*}{\rotatebox[origin=c]{90}{WILDS}}} & iwildcam & \underline{1.4{\small $\pm$ 1.7}} & 3.0{\small $\pm$ 0.0} & 3.2{\small $\pm$ 3.0} & 2.8{\small $\pm$ 2.5} & 4.1{\small $\pm$ 1.6} & \cellcolor{gray!15}\textbf{0.9{\small $\pm$ 0.0}} \\ 
& camelyon & \textbf{0.5{\small $\pm$ 0.9}} & 3.3{\small $\pm$ 0.7} & 0.9{\small $\pm$ 1.5} & 1.4{\small $\pm$ 1.5} & 1.2{\small $\pm$ 0.9} & \cellcolor{gray!15}\underline{0.6{\small $\pm$ 0.0}} \\ 
& fmow & 1.4{\small $\pm$ 0.9} & 1.0{\small $\pm$ 0.8} & \underline{0.7{\small $\pm$ 0.3}} & 1.2{\small $\pm$ 0.5} & 1.5{\small $\pm$ 1.4} & \cellcolor{gray!15}\textbf{0.4{\small $\pm$ 0.0}} \\ 
& civilcomments & 0.7{\small $\pm$ 1.2} & \textbf{0.0{\small $\pm$ 0.0}} & 0.3{\small $\pm$ 0.4} & \underline{0.2{\small $\pm$ 0.4}} & 0.3{\small $\pm$ 0.2} & \cellcolor{gray!15}2.8{\small $\pm$ 0.0} \\ 
\midrule
{\multirow{3}{*}{\rotatebox[origin=c]{90}{MSV}}} & cifar10-low & \underline{1.2{\small $\pm$ 1.3}} & 5.7{\small $\pm$ 0.0} & 2.8{\small $\pm$ 2.9} & 3.2{\small $\pm$ 1.2} & 6.3{\small $\pm$ 1.9} & \cellcolor{gray!15}\textbf{0.0{\small $\pm$ 0.0}} \\ 
& cifar10-high & 1.0{\small $\pm$ 1.4} & 3.2{\small $\pm$ 0.0} & 0.3{\small $\pm$ 0.2} & 1.9{\small $\pm$ 1.6} & \textbf{0.0{\small $\pm$ 0.0}} & \cellcolor{gray!15}\textbf{0.0{\small $\pm$ 0.0}} \\ 
& pacs & 1.1{\small $\pm$ 0.7} & \textbf{0.0{\small $\pm$ 0.0}} & 0.4{\small $\pm$ 0.6} & 0.6{\small $\pm$ 0.5} & 0.8{\small $\pm$ 0.5} & \cellcolor{gray!15}\textbf{0.0{\small $\pm$ 0.0}} \\ 
\midrule
{\multirow{7}{*}{\rotatebox[origin=c]{90}{GLUE}}} & cola & 2.9{\small $\pm$ 1.8} & \textbf{0.2{\small $\pm$ 0.5}} & 1.4{\small $\pm$ 1.2} & 2.8{\small $\pm$ 2.7} & \underline{0.4{\small $\pm$ 0.0}} & \cellcolor{gray!15}56.0{\small $\pm$ 0.0} \\ 
& mnli & 0.3{\small $\pm$ 0.3} & \textbf{0.1{\small $\pm$ 0.0}} & 0.4{\small $\pm$ 0.2} & 0.7{\small $\pm$ 0.8} & \underline{0.2{\small $\pm$ 0.1}} & \cellcolor{gray!15}\underline{0.2{\small $\pm$ 0.0}} \\ 
& qnli & 0.7{\small $\pm$ 1.1} & \textbf{0.0{\small $\pm$ 0.0}} & 1.6{\small $\pm$ 2.1} & 1.5{\small $\pm$ 1.9} & \underline{0.1{\small $\pm$ 0.2}} & \cellcolor{gray!15}0.4{\small $\pm$ 0.0} \\ 
& qqp & 1.0{\small $\pm$ 0.6} & \underline{0.2{\small $\pm$ 0.3}} & 1.0{\small $\pm$ 1.1} & 0.9{\small $\pm$ 0.7} & 0.7{\small $\pm$ 0.9} & \cellcolor{gray!15}\textbf{0.0{\small $\pm$ 0.0}} \\ 
& rte & 1.2{\small $\pm$ 2.6} & \textbf{0.0{\small $\pm$ 0.0}} & 0.8{\small $\pm$ 1.8} & 4.3{\small $\pm$ 4.2} & \underline{0.0{\small $\pm$ 0.0}} & \cellcolor{gray!15}\textbf{0.0{\small $\pm$ 0.0}} \\
& sst2 & 0.8{\small $\pm$ 0.7} & 0.1{\small $\pm$ 0.0} & 1.0{\small $\pm$ 0.9} & 2.0{\small $\pm$ 2.9} & \textbf{0.0{\small $\pm$ 0.0}} & \cellcolor{gray!15}\textbf{0.0{\small $\pm$ 0.0}} \\ 
& mrpc & \underline{0.2{\small $\pm$ 0.2}} & 0.2{\small $\pm$ 0.1} & 0.4{\small $\pm$ 0.5} & 2.6{\small $\pm$ 2.9} & \textbf{0.0{\small $\pm$ 0.0}} & \cellcolor{gray!15}0.5{\small $\pm$ 0.0} \\ 
\bottomrule
\end{tabular}

}

\caption{\textbf{Average instantaneous regret at step 100.} Mean and standard deviation over 5 random seeds.}
\label{tab:instant100}
\end{table}

\subsubsection{Success rate}

\begin{figure}[h!]
    \centering
    \includegraphics[width=\linewidth]{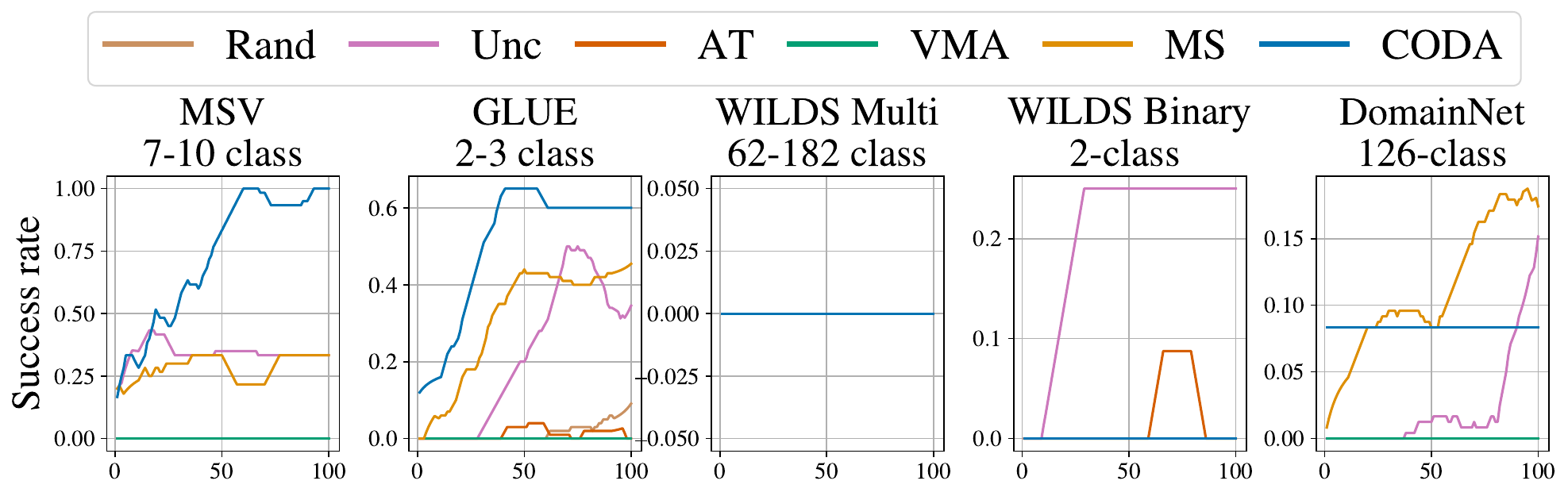}
    \caption{\textbf{``Success rate'' of each model in selecting the \textit{absolute best} model at each time step.} Mean over 5 random seeds. In all datasets, several methods have not yet selected the absolute best model by time step 100.}
    \label{fig:successrate}
\end{figure}

\subsection{Unsupervised model selection results}

\begin{table}[h]
\centering
\resizebox{\linewidth}{!}{
\begin{tabular}{llrrrrrr}
\toprule
& Task & Source val & DEV & Entropy & BNM & EnsV & \cellcolor{gray!15}CODA \\
\midrule

{\multirow{12}{*}{\rotatebox[origin=c]{90}{DomainNet126}}} & real $\rightarrow$ sketch & 0.5 & 4.6 & 3.1 & 3.1 & \textbf{0.4} & \cellcolor{gray!15}{1.0} \\
& real $\rightarrow$ clip & 4.5 & 49.3 & 0.3 & 6.5 & {2.8} & \cellcolor{gray!15}\textbf{2.3} \\
& real $\rightarrow$ paint & 1.3 & 34.7 & 2.3 & 2.0 & \textbf{0.2} & \cellcolor{gray!15}{1.2} \\
& sketch $\rightarrow$ real & 4.7 & 6.3 & 9.1 & 8.5 & {4.5} & \cellcolor{gray!15}\textbf{0.1} \\
& sketch $\rightarrow$ clip & 2.4 & 13.4 & 5.9 & 6.4 & \textbf{3.0} & \cellcolor{gray!15}{4.8} \\
& sketch $\rightarrow$ paint & 3.8 & 3.5 & 3.2 & 3.2 & {1.7} & \cellcolor{gray!15}\textbf{0.1} \\
& clip $\rightarrow$ real & 1.5 & 6.3 & 5.6 & 5.6 & {0.5} & \cellcolor{gray!15}\textbf{0.3} \\
& clip $\rightarrow$ sketch & 2.8 & 4.9 & 5.1 & 4.9 & {2.2} & \cellcolor{gray!15}\textbf{0.4} \\
& clip $\rightarrow$ paint & 3.1 & 7.8 & 4.2 & 4.2 & {4.2} & \cellcolor{gray!15}\textbf{1.2} \\
& paint $\rightarrow$ real & 0.0 & 36.6 & 3.3 & {3.3} & \textbf{0.1} & \cellcolor{gray!15}\textbf{0.1} \\
& paint $\rightarrow$ sketch & 0.7 & 26.1 & 4.3 & 4.3 & {3.3} & \cellcolor{gray!15}\textbf{0.9} \\
& paint $\rightarrow$ clip & 2.9 & 16.2 & 6.3 & 6.3 & {1.1} & \cellcolor{gray!15}\textbf{0.4} \\

\midrule

{\multirow{4}{*}{\rotatebox[origin=c]{90}{WILDS}}} & iwildcam & 9.8 & - & - & - & \textbf{0.9} & \cellcolor{gray!15}{6.1} \\
& camelyon & \textbf{4.3} & - & - & - & 13.1 & \cellcolor{gray!15}{11.3} \\
& fmow & {0.8} & - & - & - & 0.9 & \cellcolor{gray!15}\textbf{0.8} \\
& civilcomments & - & - & - & - & \textbf{3.8} & \cellcolor{gray!15}{4.7} \\

\midrule

{\multirow{3}{*}{\rotatebox[origin=c]{90}{MSV}}} & {cifar10-low} & - & - & - & - & \textbf{2.3} & \cellcolor{gray!15}{\textbf{2.3}} \\
& {cifar10-high} & - & - & - & - & \textbf{4.9} & \cellcolor{gray!15}{\textbf{4.9}} \\
& {pacs} & - & - & - & - & \textbf{0.4} & \cellcolor{gray!15}\textbf{\textbf{0.4}} \\

\midrule

{\multirow{7}{*}{\rotatebox[origin=c]{90}{GLUE}}} & {cola} & - & - & - & - & 5.3 & \cellcolor{gray!15}{\textbf{5.0}} \\
& {mnli} & - & - & - & - & \textbf{3.0} & \cellcolor{gray!15}{\textbf{3.0}} \\
&  {qnli}  & - & - & - & - & \textbf{3.3} & \cellcolor{gray!15}\textbf{ \textbf{3.3}} \\
&  {qqp}  & - & - & - & - & \textbf{1.1} & \cellcolor{gray!15}\textbf{ \textbf{1.1}} \\
&  {rte}  & - & - & - & - & \textbf{14.8} & \cellcolor{gray!15}\textbf{\textbf{14.8}} \\
&  {sst2}  & - & - & - & - & \textbf{3.6} & \cellcolor{gray!15}\textbf{ \textbf{3.6}} \\
&  {mrcp}  & - & - & - & - & \textbf{1.0} & \cellcolor{gray!15}{\textbf{1.0}} \\

\bottomrule
\end{tabular}
}
\caption{\textbf{Unsupervised model selection results.} We report regret at step 0 (lower is better) for all methods on all tasks. Best method for each task is in bold. CODA matches or exceeds state-of-the-art performance on 20 out of 26 tasks. Note that because models/predictions for MSV and GLUE are black-box/one-hot (as in ModelSelector~\cite{okanovic2024all}), the only comparison we are able to make is to EnsV.}
\vspace{-10pt}
\end{table}

Our method defines a prior over model performance that creates a strong starting point for active model selection. This can be used in isolation, without active label collection, to perform unsupervised model selection. We compare against five existing methods for model selection in unsupervised domain adaptation:

\noindent \textbf{Source validation} Model selection is performed using validation accuracy on ``source'' data, \ie using a validation set from the same distribution as the training set.

\noindent \textbf{Target entropy~\cite{musgrave2022three}} Shannon entropy is computed on prediction scores on the test set. Model selection is performed by selecting the model with the lowest entopy, \ie the model that is most confident in its test predictions.

\noindent \textbf{Deep embedded validation~\cite{you2019towards}} Computes a classification loss for each source validation sample, and weights each loss based on a computed probability that the sample ``belongs to'' the test domain. This probability comes from a separate domain classifier trained on source and test data.

\noindent \textbf{Batch nuclear norm~\cite{cui2020towards}} Performs singular value decomposition on the prediction matrix. Model selection is performed by selecting the model with the minmium nuclear norm (\ie minimum sum of singular values).

\noindent \textbf{EnsV~\cite{hu2025towards}} All models in the hypothesis set are ensembled. To perform model selection, accuracy is estimated for each model with respect to the ensemble's predictions.


\subsection{Unaggregated results on all tasks}

\begin{figure*}
    \centering
    \includegraphics[width=0.85\linewidth]{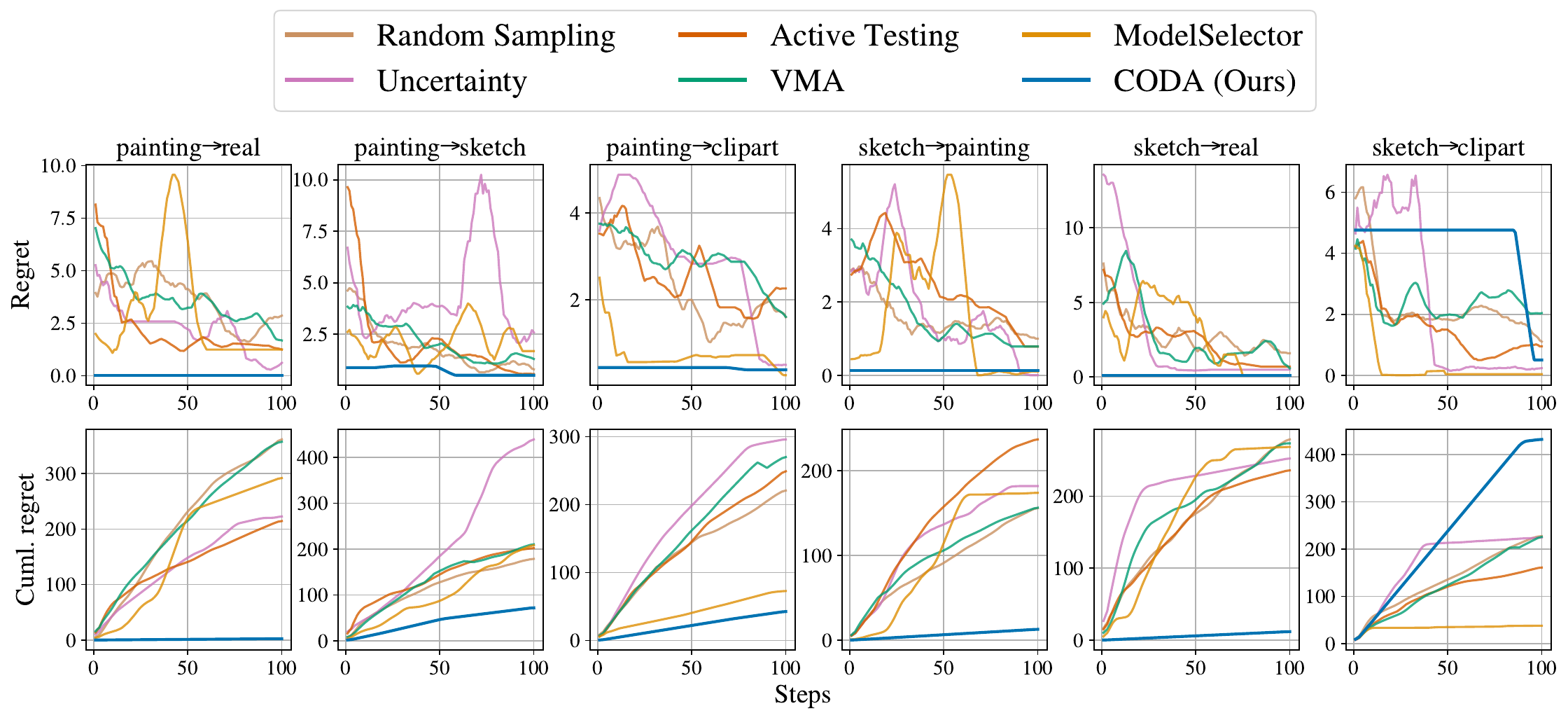}
    \includegraphics[width=0.85\linewidth]{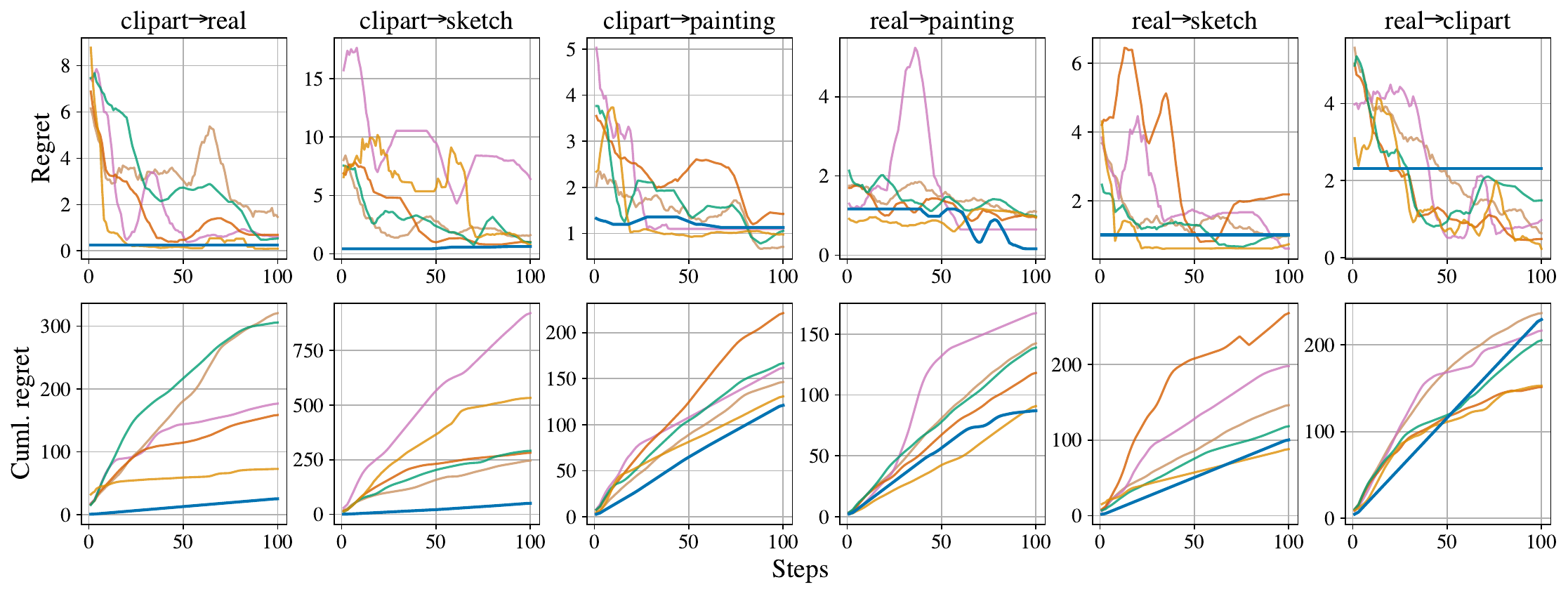}
    \includegraphics[width=0.85\linewidth]{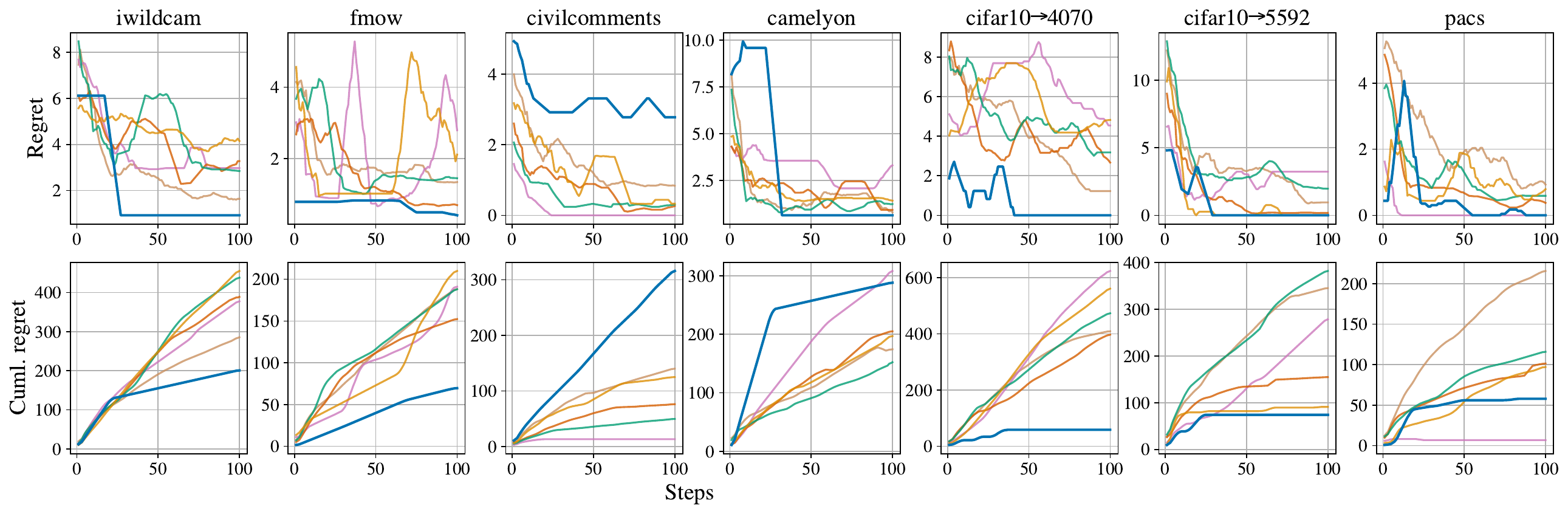}
    \includegraphics[width=0.85\linewidth]{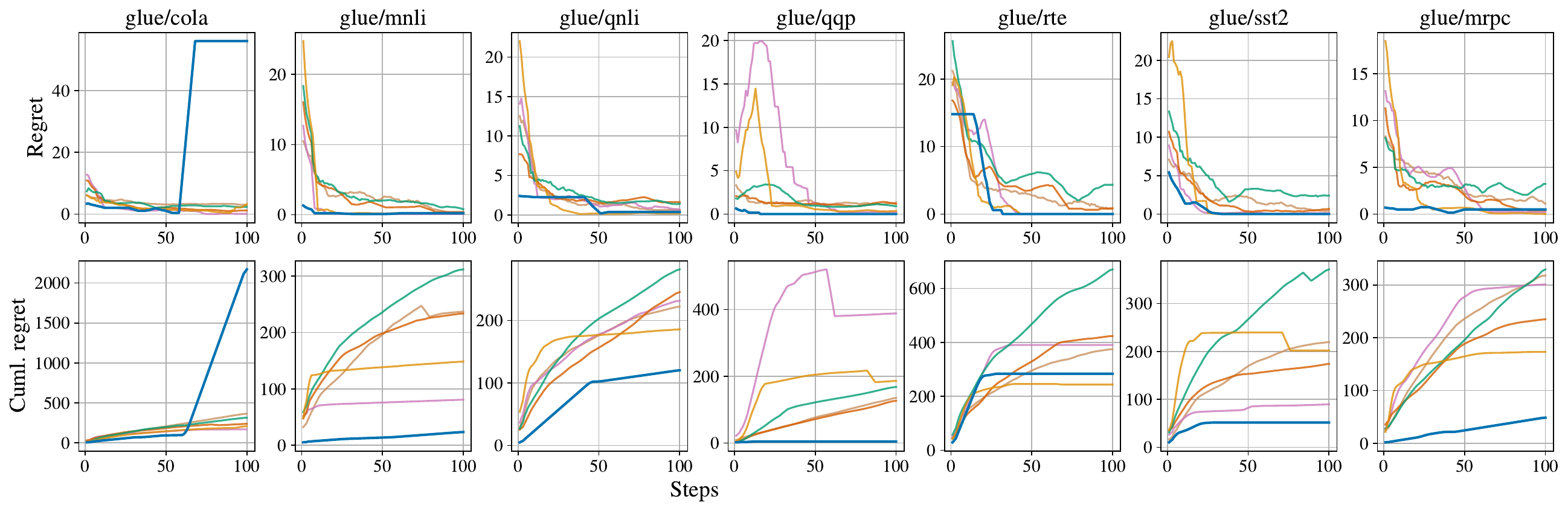}
    \vspace{-10pt}
    \caption{Results on all benchmarks.}
    \label{fig:all-results}
\end{figure*}

In \cref{fig:all-results} we visualize regret and cumulative regret at every step for all baselines on all tasks. 



\clearpage

\section{Implementation details}

\subsection{Dawid-Skene data generating process} 

\begin{figure}[h!]
    \centering
    \includegraphics[width=0.8\linewidth]{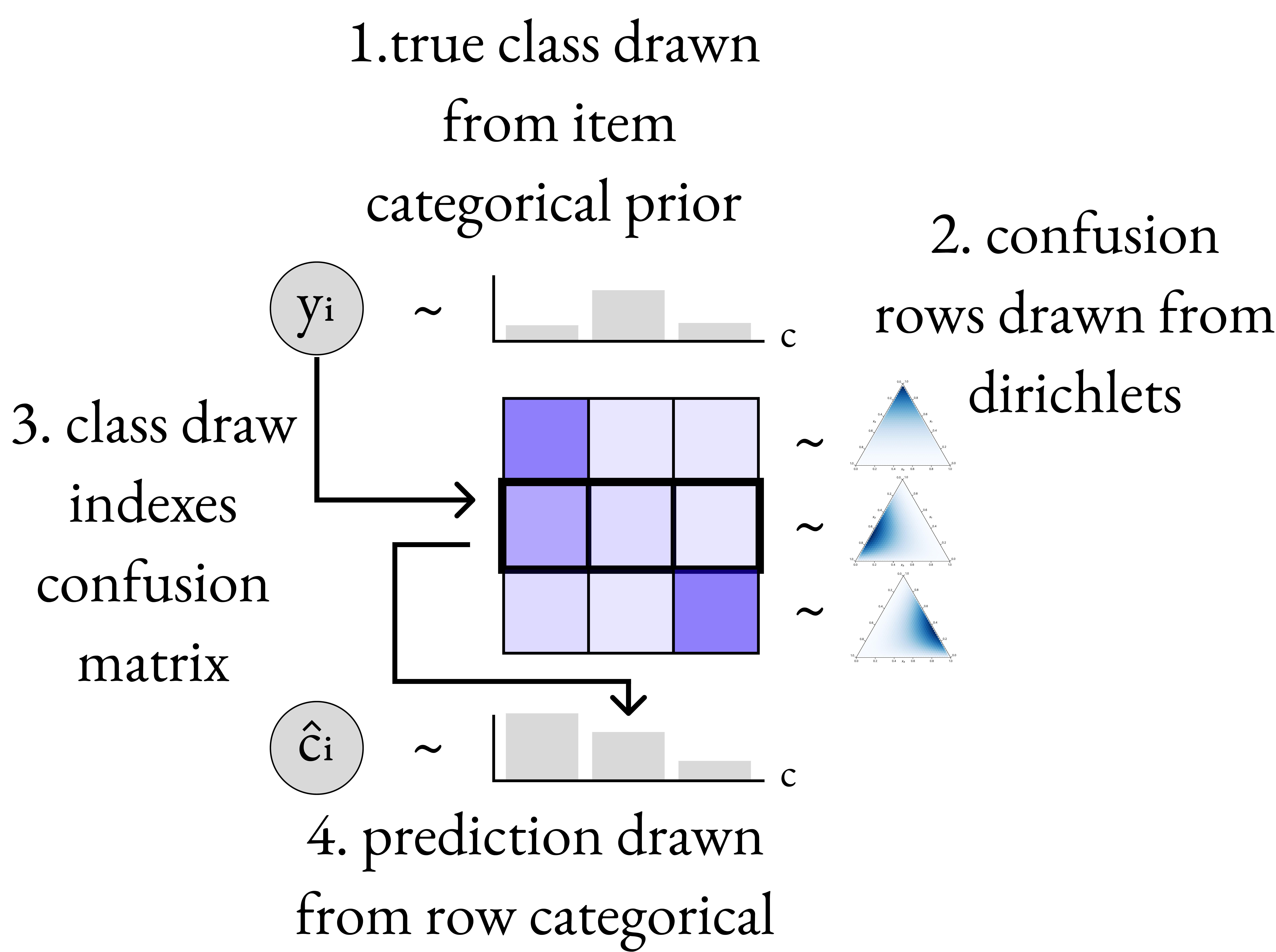}
    \caption{A visual depiction of the Dawid-Skene~\cite{dawid1979maximum} data generating process that we adapt to active model selection. See \cref{sec:ds} of the main paper for more details.}
    \label{fig:ds}
\end{figure}

We visualize the data generating process of the Bayesian implementation of the Dawid-Skene model in \cref{fig:ds}, as described in \cref{sec:ds}. We repeat the text here for easy reference.

The data generating process proceeds as follows:
\begin{enumerate}
    \item Each data point's true class label $y_i$ is drawn randomly from per-data-point prior distributions over which class that data point could be, $y_i \sim \text{Cat}(\pi(x_i))$. 
    
    \item Each row of the classifier's confusion matrix is drawn randomly from per-row distributions, $M_{k,\, c,\cdot} \sim \theta_{k,c}$, where $\theta_{k,c}$ is the prior distribution over what the row of the confusion matrix could be. To accommodate Bayesian updates, we initialize each $\theta_{k,c}$ to be a Dirichlet prior.
    \item The sampled true class indexes into the corresponding row of the classifier's confusion matrix, $M_{k,\, y_{i}}$.
    \item The classifier's prediction for that data point is sampled from the distribution over that row's cells, ${\hat c_{k, i} \sim \text{Cat}(M_{k,\, y_{i}}})$. 
\end{enumerate}

\begin{figure}[h!]
    \centering
    \includegraphics[width=0.8\linewidth]{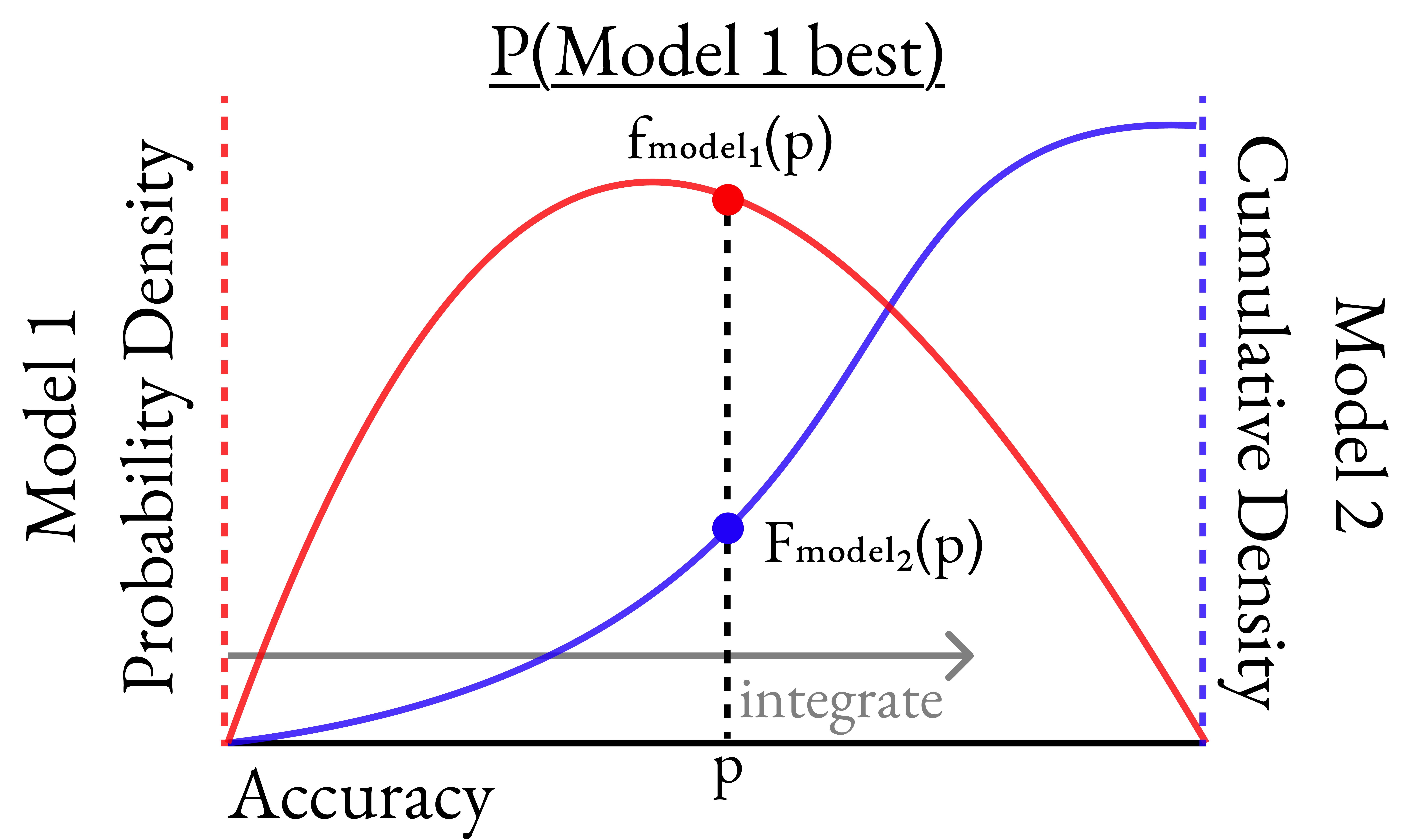}
    \caption{A visual depiction of the integration technique used to construct our distribution over which model is best, \pbest. See \cref{sec:pbest_supp} of the supplemental and \cref{sec:pbest} of the main paper for more details.}
    \label{fig:pbest}
\end{figure}

\subsection{Computing \pbest}
\label{sec:pbest_supp}

We illustrate visually the computation of \pbest from \cref{sec:pbest} in \cref{fig:pbest} in the simplified case of two models. To compute the probability that Model 1 is best, we integrate over all possible accuracy values that Model 1 \textit{could} have. For every possible accuracy, we compute the probability that Model 1 has that accuracy (defined by its PDF $f$), multiplied that the probability Model 2 has accuracy \textit{less} than that value (defined by its CDF $F$).

\section{Data and model details}

We provide more details about the datasets and models in our benchmarking suite in \cref{tab:domainnet}, \cref{tab:wilds}, and \cref{tab:msv}.

\begin{table}[h!]
\resizebox{\linewidth}{!}{
\begin{tabular}{lrllr}

\toprule

\textbf{DomainNet126}              & \multicolumn{1}{l}{}             &            &                         & \multicolumn{1}{l}{}              \\
Task                            & \multicolumn{1}{l}{Num. Classes} & \multicolumn{2}{c}{Num. Checkpoints} & \multicolumn{1}{l}{Test set size} \\

\midrule

\multirow{2}{*}{sketch\_painting}  & \multirow{2}{*}{126}             & 10 algs    & \multirow{2}{*}{= 200}  & \multirow{2}{*}{3086}             \\

                                   &                                  & 20 epochs  &                         &                                   \\
                                   \midrule
\multirow{2}{*}{sketch\_real}      & \multirow{2}{*}{126}             & 10 algs    & \multirow{2}{*}{= 200}  & \multirow{2}{*}{20939}            \\
                                   &                                  & 20 epochs  &                         &                                   \\
                                   \midrule
\multirow{2}{*}{sketch\_clipart}   & \multirow{2}{*}{126}             & 10 algs    & \multirow{2}{*}{= 200}  & \multirow{2}{*}{5611}             \\
                                   &                                  & 20 epochs  &                         &                                   \\
                                   \midrule
\multirow{2}{*}{clipart\_painting} & \multirow{2}{*}{126}             & 10 algs    & \multirow{2}{*}{= 200}  & \multirow{2}{*}{3086}             \\
                                   &                                  & 20 epochs  &                         &                                   \\
                                   \midrule
\multirow{2}{*}{clipart\_sketch}   & \multirow{2}{*}{126}             & 10 algs    & \multirow{2}{*}{= 200}  & \multirow{2}{*}{7313}             \\
                                   &                                  & 20 epochs  &                         &                                   \\
                                   \midrule
\multirow{2}{*}{clipart\_real}     & \multirow{2}{*}{126}             & 10 algs    & \multirow{2}{*}{= 200}  & \multirow{2}{*}{20939}            \\
                                   &                                  & 20 epochs  &                         &                                   \\
                                   \midrule
\multirow{2}{*}{painting\_sketch}  & \multirow{2}{*}{126}             & 10 algs    & \multirow{2}{*}{= 200}  & \multirow{2}{*}{7313}             \\
                                   &                                  & 20 epochs  &                         &                                   \\
                                   \midrule
\multirow{2}{*}{painting\_real}    & \multirow{2}{*}{126}             & 10 algs    & \multirow{2}{*}{= 200}  & \multirow{2}{*}{20939}            \\
                                   &                                  & 20 epochs  &                         &                                   \\
                                   \midrule
\multirow{2}{*}{painting\_clipart} & \multirow{2}{*}{126}             & 10 algs    & \multirow{2}{*}{= 200}  & \multirow{2}{*}{5611}             \\
                                   &                                  & 20 epochs  &                         &                                   \\
                                   \midrule
\multirow{2}{*}{real\_painting}    & \multirow{2}{*}{126}             & 10 algs    & \multirow{2}{*}{= 200}  & \multirow{2}{*}{3086}             \\
                                   &                                  & 20 epochs  &                         &                                   \\
                                   \midrule
\multirow{2}{*}{real\_sketch}      & \multirow{2}{*}{126}             & 10 algs    & \multirow{2}{*}{= 200}  & \multirow{2}{*}{7313}             \\
                                   &                                  & 20 epochs  &                         &                                   \\
                                   \midrule
\multirow{2}{*}{real\_clipart}     & \multirow{2}{*}{126}             & 10 algs    & \multirow{2}{*}{= 200}  & \multirow{2}{*}{5611}             \\
                                   &                                  & 20 epochs  &                         &                                  \\
\bottomrule
\end{tabular}
}
\caption{\textbf{Dataset details: DomainNet126.} For each transfer task, we first train a ``source-only'' model on the source domain. We then train 10 UDA models for each transfer task (source domain $\rightarrow$ target domain) using the Powerful Benchmarker codebase~\cite{musgrave2022three}.}
\label{tab:domainnet}
\end{table}

\begin{table}[]
\resizebox{\linewidth}{!}{
\begin{tabular}{lrllrr}
\toprule
\textbf{WILDS} & \multicolumn{1}{l}{} &  &  & \multicolumn{1}{l}{} &  \\
Task & \multicolumn{1}{l}{Num. Classes} & \multicolumn{2}{c}{Num. Checkpoints} & \multicolumn{1}{l}{Test set size} & Regret w/ val. \\
\midrule

\multirow{2}{*}{RxRx1} & \multirow{2}{*}{1139} & 4 algs & \multirow{2}{*}{= 72} & \multirow{2}{*}{34,432} & \multicolumn{1}{r}{0.1} \\
 &  & 18 epochs &  &  &  \\
 \midrule

\multirow{2}{*}{Amazon} & \multirow{2}{*}{5} & 4 algs & \multirow{2}{*}{= 12} & \multirow{2}{*}{100,050} & \multicolumn{1}{r}{0.1} \\
 &  & 3 epochs &  &  &  \\
 \midrule

\multirow{2}{*}{CivilComments} & \multirow{2}{*}{2} & 4 algs & \multirow{2}{*}{= 20} & \multirow{2}{*}{133,782} & N/A \\
 &  & 5 epochs &  &  &  \\
 \midrule

\multirow{2}{*}{fMoW} & \multirow{2}{*}{62} & 4 algs & \multirow{2}{*}{= 240} & \multirow{2}{*}{22,108} & \multicolumn{1}{r}{0.8} \\
 &  & 60 epochs &  &  &  \\
 \midrule

\multirow{2}{*}{iWildCam} & \multirow{2}{*}{182} & 4 algs & \multirow{2}{*}{= 48} & \multirow{2}{*}{42,791} & \multicolumn{1}{r}{9.8} \\
 &  & 12 epochs &  &  &  \\
 \midrule

\multirow{2}{*}{Camelyon17} & \multirow{2}{*}{2} & 4 algs & \multirow{2}{*}{= 40} & \multirow{2}{*}{85,054} & \multicolumn{1}{r}{4.3} \\
 &  & 10 epochs &  &  & \\
 \bottomrule

\end{tabular}
}
\caption{\textbf{Dataset details: WILDS.} 
We show metrics for all classification datasets in WILDS where we could perform model selection. In our main experiments, we only use the benchmarks where near-perfect model selection can be performed trivially using the default in-distribution validation set. We train all models ourselves using the public code from \citet{koh2021wilds}.
}
\label{tab:wilds}
\end{table}

\begin{table}[]
\resizebox{\linewidth}{!}{
\begin{tabular}{lrlrr}
\toprule
\textbf{MSV} & \multicolumn{1}{l}{} &  & \multicolumn{1}{l}{} & \multicolumn{1}{l}{} \\
Dataset & \multicolumn{1}{l}{Num. Classes} & \multicolumn{2}{c}{Num. Checkpoints} & \multicolumn{1}{l}{Test set size} \\
CIFAR10-High & 10 &  & 80 & 10000 \\
CIFAR10-Low & 10 &  & 80 & 10000 \\
PACS & 7 &  & 30 & 9991 \\
\bottomrule
 & \multicolumn{1}{l}{} &  & \multicolumn{1}{l}{} & \multicolumn{1}{l}{} \\
\toprule
\textbf{GLUE} & \multicolumn{1}{l}{} &  & \multicolumn{1}{l}{} & \multicolumn{1}{l}{} \\
Dataset & \multicolumn{1}{l}{Num. Classes} & \multicolumn{2}{c}{Num. Checkpoints} & \multicolumn{1}{l}{Test set size} \\
CoLA & 2 &  & 109 & 1043 \\
MNLI & 3 &  & 82 & 9815 \\
QNLI & 2 &  & 90 & 5463 \\
QQP & 2 &  & 101 & 40430 \\
RTE & 2 &  & 87 & 277 \\
SST2 & 2 &  & 97 & 872 \\
MRPC & 2 &  & 95 & 408 \\
\bottomrule
\end{tabular}
}
\caption{\textbf{Dataset details: MSV and GLUE.} All model checkpoints sourced directly from \citet{okanovic2024all}.}
\label{tab:msv}
\end{table}

\end{document}